\documentclass[10pt,twocolumn,letterpaper]{article}
\usepackage[pagenumbers]{cvpr}

\usepackage{graphicx}
\usepackage{amsmath}
\usepackage{amssymb}
\usepackage{booktabs}

\usepackage{kotex}
\usepackage{hhline}
\usepackage{makecell, multirow}
\usepackage{subcaption}
\usepackage{array}

\usepackage[pagebackref,breaklinks,colorlinks]{hyperref}

\usepackage[capitalize]{cleveref}
\crefname{section}{Sec.}{Secs.}
\Crefname{section}{Section}{Sections}
\Crefname{table}{Table}{Tables}
\crefname{table}{Tab.}{Tabs.}

\let\svthefootnote\thefootnote
\newcommand\freefootnote[1]{%
  \let\thefootnote\relax%
  \footnotetext{#1}%
  \let\thefootnote\svthefootnote%
}

\begin{document}

\title{Deep Collective Knowledge Distillation}

\author{
   Jihyeon Seo$^*$\\
   Samsung SDS\\
   {\tt\small jihyeon0.seo@samsung.com}\\
   \and
   Kyusam Oh$^{*,\dag}$\\
   SK ecoplant\\
   {\tt\small q3.oh@sk.com}\\
   \and
   Chanho Min$^\dag$\\
   Ajou University\\
   {\tt\small chanhomin@ajou.ac.kr}\\
   \and
   Yongkeun Yun$^\dag$\\
   SK ecoplant\\
   {\tt\small yongkeun.yun@sk.com}\\
   {\tt\small akanova@korea.ac.kr}\\
   \and
   Sungwoo Cho\\
   Samsung SDS\\
   {\tt\small sung-woo.cho@samsung.com}\\
}
\maketitle

\begin{abstract}
	Many existing studies on knowledge distillation have focused on methods in which a student model mimics a teacher model well.
	Simply imitating the teacher's knowledge, however, is not sufficient for the student to surpass that of the teacher.
	We explore a method to harness the knowledge of other students to complement the knowledge of the teacher.  
	We propose deep collective knowledge distillation for model compression, called DCKD, which is a method for training student models with rich information to acquire knowledge from not only their teacher model but also other student models. 
	The knowledge collected from several student models consists of a wealth of information about the correlation between classes. 
	Our DCKD considers how to increase the correlation knowledge of classes during training. 
	Our novel method enables us to create better performing student models for collecting knowledge.
	This simple yet powerful method achieves state-of-the-art performances in many experiments.
	For example, for ImageNet, ResNet18 trained with DCKD achieves 72.27\%, which outperforms the pretrained ResNet18 by 2.52\%.
	For CIFAR-100, the student model of ShuffleNetV1 with DCKD achieves 6.55\% higher top-1 accuracy than the pretrained ShuffleNetV1.
\end{abstract}

\freefootnote{*These authors contributed equally to this work.}
\freefootnote{\dag These authors conducted this research while they worked at Samsung SDS.}
\section{Introduction}
\label{sec:intro}

Knowledge distillation \cite{KD} is an effective method for compressing a heavy teacher model to a lighter student model. 
The main idea of knowledge distillation is that a teacher model with higher capacity and better performance distills the softened output distribution into a student model as knowledge. 
Hinton \etal \cite{KD} explained that softened outputs with higher entropy than hard labels provide much richer information. 
In previous studies \cite{model_compression, KD, entropy}, assigning the probabilities to other classes, which leads to increased entropy, was effective in generalizing a network. 
These probabilities provided valuable information about the correlation between other classes and Dubey \etal \cite{max-entropy} showed that maximum entropy training with these correlation information is effective.
Many studies \cite{FitNet, FT, CC, reviewkd, overhaul, RKD, CRD, AT} on knowledge distillation have focused on efficiently transferring teacher’s knowledge to students.
We took a step further by adopting a different approach to knowledge enrichment.
If a student model learns from only a teacher’s soft targets, then the student will only imitate the teacher. 
However, if the knowledge of peer students is additionally given, it can help a student model outperform the student model learned only from a teacher.
Our work focuses on creating a student model that can have rich representation by training not only with the knowledge provided by the teacher but also with additional knowledge of various correlations between classes provided by the peer students.
As a method for generating additional knowledge, we propose gathering knowledge from multiple students.

Mutual learning methods \cite{OKDDIP, KDCL, ONE, DML} aim to train powerful student models using ensembled knowledge of multiple untrained students without a pretrained teacher model. 
These methods train multistudent models with knowledge generated on-the-fly from students’ logits, and this online knowledge is often generated to better represent the ground truth or soft targets of the teacher model.
However, we propose an approach for generating additional online knowledge containing diverse correlation information from multistudent models, not similar to the ground truth or the teacher's soft targets.
Since the teacher model learned with the supervised learning manner becomes over-confident \cite{mixup_training}, it may overlook other correlation information.
Thus, we believe that the additional knowledge, including correlational information between classes, can assist the teacher model.

We present deep collective knowledge distillation (DCKD) to improve the performance of a student model with a wealth of knowledge.
Our method achieves state-of-the-art performances in every experiment. We perform extensive experiments on KD benchmarks on several datasets, including ImageNet \cite{imagenet}, CIFAR-100, CIFAR-10 \cite{cifar} and Fashion-MNIST \cite{FMNIST}.
For CIFAR-100, the student model of ResNet32 with DCKD achieves 3.83\% higher top-1 accuracy than the pretrained ResNet32.
Our two main contributions to this research are as follows:
$ 1) $ We design a novel method for constructing the additional knowledge collection with more correlation information between classes.
$ 2) $ We analyze and modify the collection loss between each student and knowledge collection by reversing the direction of Kullback-Leibler divergence.

\section{Related Works}
{\bf Knowledge Distillation} 
The work of Hinton \etal \cite{KD} has revolutionized the literature of model compression by distilling the knowledge of a large model to a smaller model to construct a compact model. 
Hinton \etal \cite{KD} explained that the distribution of a teacher network, which is smoothed using temperature, can more clearly represent secondary probabilities. 
These secondary probabilities convey information about the correlation between labels and are crucial for teaching student networks. 
AT \cite{AT} attempted to improve knowledge distillation by including the effects of attention in models. 
FitNet \cite{FitNet} expected to mimic the teacher model by transferring the intermediate representation to the student model. 
FT \cite{FT} used paraphrased factors compressed from the teacher’s features to train the student, and RKD \cite{RKD} utilized structural relations of data.
OFD \cite{overhaul} proposed a feature distillation method, including a feature transform to keep the information of the teacher's features. 
CC \cite{CC} focused on the correlation between the instances to transfer, and CRD \cite{CRD} demonstrated how to make the student's representation more similar to the teacher's representation using a contrastive objective.
ReviewKD \cite{reviewkd} proposed a review mechanism that transfers multi-level knowledge of the teacher to one-level of the student.

{\bf Mutual Learning} We provide DCKD as a novel approach for knowledge distillation, but our work is also related to mutual learning methods because multiple student models train each other.
Deep mutual learning (DML) \cite{DML} trained multiple initialized models to gain remarkable performance without any pretrained teacher.
ONE \cite{ONE} ensembled logits of multiple students to generate an on-the-fly teacher and trained each student with it.
OKDDip \cite{OKDDIP} similarly constructed an ensemble with students but considered attention-based weights for each student.
KDCL \cite{KDCL} regarded all networks as students, proposed methods to generate a soft target from all these students, and mentioned that the generated soft target is transferred to all students as knowledge. 
The existing methods \cite{ONE, KDCL} ensemble the outputs of each student to generate a single soft target and use it to train the students with the forward direction of Kullback-Leibler divergence.
However, our method differs from these methods because our method trains each student with several different soft targets generated from different peers. 
Each student has more diversity by learning other students’ collective knowledge that does not include the individual knowledge of each student.

\begin{figure}
	\centering
	\begin{subfigure}{0.45\columnwidth}
		\centering
		\includegraphics[page=1, trim ={0cm 11.5cm 20cm 0cm}, clip, width=1\linewidth]{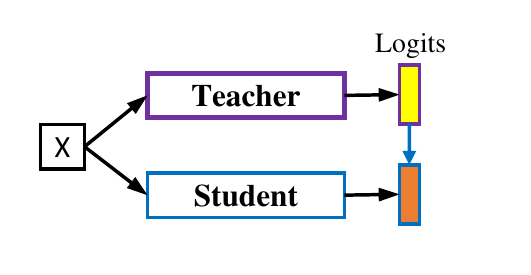}
		\caption{KD}
		\label{fig:fig1_a}
	\end{subfigure}%
	\hspace*{\fill}
	\begin{subfigure}{0.55\columnwidth}
	\centering
\includegraphics[page=2, trim ={0.5cm 11.5cm 18cm 0cm}, clip, width=1\linewidth]{figures/fig1_a.pdf}
\end{subfigure}%
		\newline
	\newline
	\begin{subfigure}{0.45\columnwidth}
		\centering
	\includegraphics[page=1, trim ={0cm 11cm 20cm 0cm}, clip, width=1\linewidth]{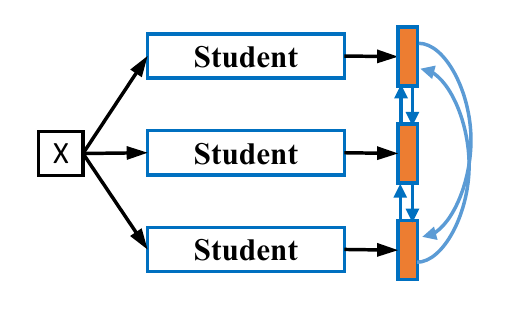}
		\caption{DML}
		\label{fig:fig1_b}
	\end{subfigure}%
\hspace*{\fill}
\begin{subfigure}{0.55\columnwidth}
	\centering
	\vspace{-5mm}
	\includegraphics[page=1, trim ={0cm 9cm 18cm 0cm}, clip, width=1\linewidth]{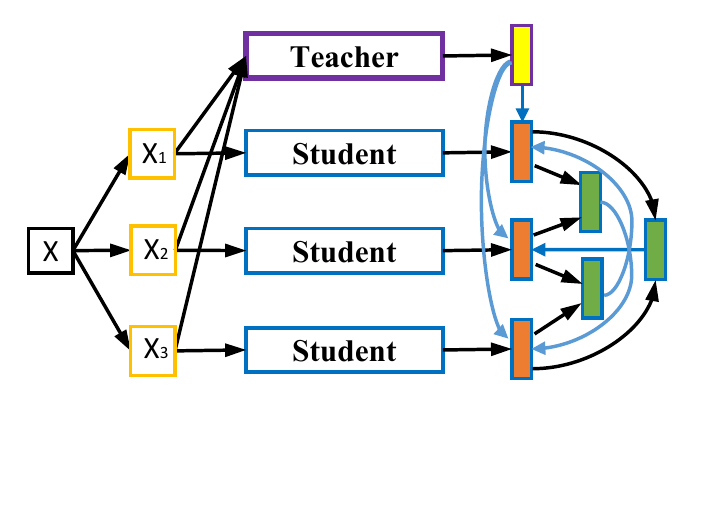}
	\vspace{-10mm}
	\caption{DCKD}
	\label{fig:fig1_c}
\end{subfigure}%


		\caption{
		(a) KD \cite{KD}: A pretrained teacher distills knowledge to a student.
		(b) DML \cite{DML}: Students learn from each other.
		(c) DCKD: A pretrained teacher distills knowledge to each student while each student learns from each knowledge collection. 
		Each knowledge collection is generated differently on-the-fly for each student model.
	}
	\label{fig:fig1}
\end{figure}
\section{Method}

\subsection{Deep Collective Knowledge Distillation}
Our method is based on the idea that a model can be a better student if the model's output represents not only the distribution of the correct class but also a relationship with other classes, as Hinton \etal described \cite{KD}.
For example, we assume that there is input data of the first class, which has features that are similar to the input data of the second class.
If a model’s output has a one-hot distribution, \eg, ${\mathbf p}=(1,0,0,\cdots, 0)$, it only has information about the correct class and does not have any knowledge concerning other classes.
On the other hand, if a model’s output distribution is ${\mathbf p}=(0.8,0.2,0,\cdots,0)$, it can be considered that the input data is an object of the first class and shares some similarities with the input data of the second class. 
The similarities among the input data between two classes can help a student better learn the features of the input data.

As shown in \cref{fig:fig1_c}, each student is trained using the knowledge distillation loss with the teacher’s knowledge and using the collection loss with the collective knowledge constructed from other students, excluding oneself.
After the training is finished, all students are independently utilized for the inference phase.

For a better mathematical definition, we denote the set of models as $F=\{f_i\}$. 
Let $\mathbf{x}$ be an input data chosen from dataset $D$ and ${\mathbf y}$ be logits of a model, \eg, ${\mathbf y}=(y^1 , \cdots, y^c)$, where $c$ is a class index.
We define logits ${\mathbf y}_i$ and probability distribution ${\mathbf p}_i$ as follows:
\begin{align}
	\mathbf{y}_i &= f_i(\mathbf{x}),\\
	\mathbf{p}_i &= h(\mathbf{y}_i).
\end{align}
We note that $f_i$ outputs logits $\mathbf{y}_i$ instead of probability distribution $\mathbf{p}_i$. $\mathbf{p}_i$ is the probability distribution of the $i$-th student, and $h$ is typical softmax activation function.

{\bf Loss function} With the definition of $\mathbf{p}_i$ and $\mathbf{y}_i$, we state the $k$-th student's loss as
\begin{align}
	\mathcal{L}_{k}(\mathbf{p}_{h},\mathbf{p}_{T},\{\mathbf{p}_i\}, \mathbf{p}_k)&=\beta_{CE}\mathcal{L}_{CE}(\mathbf{p}_{h},\mathbf{p}_k) \\
	&+\label{KD_loss}\beta_{KD}\mathcal{L}_{KD}(\mathbf{p}_T,\mathbf{p}_k) \\
	&+ \beta_{Col}\mathcal{L}_{Col}(\{\mathbf{p}_i\},\mathbf{p}_k).
\end{align}
where $\mathbf{p}_{h}$ represents a hard label, $\mathbf{p}_{T}$ is the distribution of the teacher model's output, $\{\mathbf{p}_i\}$ is the set of the distribution of all student models' outputs and $\mathbf{p}_k$ is the distribution of the $k$-th student's output. 
Each $\beta$ is a constant weight for balancing each loss.

$\mathcal{L}_{k}$ is composed of the following three parts: cross-entropy loss $\mathcal{L}_{CE}$, knowledge distillation loss $\mathcal{L}_{KD}$ and collection loss $\mathcal{L}_{Col}$.
First, $\mathcal{L}_{CE}$ is a typical loss for a classification task.
Second, the standard $\mathcal{L}_{KD}$ is defined as follows:
\begin{align}
	\mathcal{L}_{KD}(\mathbf{p}_T,\mathbf{p}_k)=
	\mathcal{L}_{CE}\left(h\left(\frac{\mathbf{y}_{T}}{T_{KD}}\right),h\left(\frac{\mathbf{y}_{k}}{T_{KD}}\right)\right),
\end{align}
where $\mathbf{y}_k$ and $\mathbf{y}_T$ are logits of $\mathbf{p}_k$ and $\mathbf{p}_T$, respectively. The temperature parameter $T_{KD}$ softens the teacher's output distribution, which is similar to the one-hot distribution, to obtain the higher entropy distribution with apparent secondary probabilities.
Last, $\mathcal{L}_{Col}$ works as a bridge between each student and each collective knowledge.
To compare two distributions, $\mathcal{L}_{Col}$ is based on the Kullback-Leibler divergence (KLD) and applies it in the reverse direction.
To effectively conserve and transfer the correlation information, we construct $\mathcal{L}_{Col}$ as follows:
\begin{align}\label{backKLD}
	\mathcal{L}_{Col}(\{\mathbf{p}_i\},\mathbf{p}_k)=\mathcal{L}_{KLD}(\hat{\mathbf{p}}_{kCol},\hat{\mathbf{p}}_k)=\mathcal{L}_{KLD}(\hat{\mathbf{p}}_k||\hat{\mathbf{p}}_{kCol}).
\end{align}
$\hat{\mathbf{p}}_{k}$ and $\hat{\mathbf{p}}_{kCol}$ are defined as follows: 
\begin{equation}
	\hat{\mathbf{p}_k}=h\left(\dfrac{\mathbf{y}_k}{T_{KLD}}\right)\quad \hat{\mathbf{p}}_{kCol}=h\left(\frac{\max  \{\mathbf{y}_i\}_{i\neq k}}{T_{KLD}}\right),
\end{equation}
where $ T_{KLD} $ is a temperature parameter for $ \mathcal{L}_{Col}$.
We will discuss $\mathcal{L}_{Col}$ more thoroughly in the following section.
%
%

Last, we state the total loss for training DCKD's students:
\begin{align}\label{Ltot}
	\mathcal{L}_{total}(\mathbf{p}_{h},\mathbf{p}_{T},\{\mathbf{p}_i\})=\sum_{i}\mathcal{L}_{i}.
\end{align}
We do not separately train the $k$-th student by calculating the gradients of each $\mathcal{L}_{i}$; rather, we simultaneously train every student with the gradients of  $\mathcal{L}_{total}$.
The main difference between the separate training and the simultaneous training is that the simultaneous training optimizes both distributions in $\mathcal{L}_{KLD}(\hat{\mathbf{p}}_k||\hat{\mathbf{p}}_{kCol})$, while the separate training optimizes $\hat{\mathbf{p}}_{k}$ only.

\newcolumntype{?}{!{\vrule width 1pt}}
\newcolumntype{P}[1]{>{\centering\arraybackslash}p{#1}}
\newcolumntype{C}{>{\centering\arraybackslash}p{1.38cm}}

{\bf Collection methods} We consider three methods for collecting knowledge to efficiently preserve the correlation between classes: logit max collection, probability max collection and average collection. 
Each collection method is utilized to find $\hat{\mathbf{p}}_{kCol}$ and is defined as follows:
\begin{itemize}
	\item Logit max collection
	\begin{equation}
		\label{log_max}\mathbf{P}_{LMax}(\{\mathbf{y}_i\}):= h(\max \{\mathbf{y}_i\}),
	\end{equation}
	\item Probability max collection: 
	\begin{equation}
		\label{p_max}\mathbf{P}_{Max}(\{\mathbf{y}_i\}):= \dfrac{\max h(\{\mathbf{y}_i\})}{\sum _{class}(\max h(\{\mathbf{y}_i\}))},
	\end{equation}
	\item Average collection:
	\begin{equation}
		\label{avg}\mathbf{P}_{avg}(\{\mathbf{y}_i\}):= avg({h(\{\mathbf{y}_i\})}).
	\end{equation}
\end{itemize}
In these equations, $avg$ and $\max$ are applied in each class index of $\{\mathbf{y}_i\}$. 

$\mathbf{P}_{Max}$ and $\mathbf{P}_{LMax}$ collect max values from each class in students' logits, whereas $\mathbf{P}_{avg}$ considers the max values as noise and cancels out by applying the average function for each class.
From our point of view, the max collection method conveys more correlations between classes, while the average collection method erases these information.
Thus, when we collect knowledge from several student models, the max collection method generates the distribution, which has rich information, while the average collection method generates the distribution, which is similar to the one-hot distribution.

The main difference between $\mathbf{P}_{Max}$ and $\mathbf{P}_{LMax}$ is whether $\max$ applies to logits $ \mathbf{y}_i $ or the distribution $ \mathbf{p}_i $. 
$\mathbf{P}_{Max}$ is equal to $\mathbf{P}_{LMax}$ if $\mathbf{y}_i$ of $\mathbf{P}_{LMax}$ is normalized to the same level by the following equation: $\mathbf{y}_i=\log(softmax(\bf{y}_i))$. 
Therefore, $\mathbf{P}_{Max}$ focuses on the relative values of probabilities, whereas $\mathbf{P}_{LMax}$ focuses on the absolute values of the logits. 
These collection methods are promising in theory, so we performed experiments to compare these methods.
Empirically, $\mathbf{P}_{LMax}$ performed better, so we adopted $\mathbf{P}_{LMax}$ to gather knowledge from multiple students.

\begin{table*}[h]
	\centering
	\resizebox{\textwidth}{!}{
		\begin{tabular}{|l|P{1.1cm}P{1.1cm}||CCCCCCC?CCC|}
			
			\hline
			\multirow{2}{*}{\textbf{Accuracy} }& \multirow{2}{*}{\textbf{Teacher}} & \multirow{2}{*}{\textbf{Student}}	
			& KD &AT &CC & OFD & CRD & CRD+KD& ReviewKD & \multicolumn{3}{c|}{\textbf{DCKD}} 	 \\
			&&& \cite{KD}&\cite{AT}& \cite{CC}&\cite{overhaul} & \cite{CRD}& \cite{CRD}& \cite{reviewkd}& \textbf{Net1}	& \textbf{Net2}& \textbf{Net3} \\
		
			\hline

\textbf{Top-1}    
& 73.31 &69.75 & \makecell{\shortstack{\rule{0in}{2ex}70.06\\\scriptsize{ ($\pm$0.11)}}}
& \makecell{\shortstack{\rule{0in}{2ex}71.55\\\scriptsize{ ($\pm$0.16)}}}     
&\makecell{\shortstack{\rule{0in}{2ex}71.45\\\scriptsize{ ($\pm$0.08)}}}
&    \makecell{\shortstack{\rule{0in}{2ex}71.35\\\scriptsize{ ($\pm$0.03)}}}
&    \makecell{\shortstack{\rule{0in}{2ex}71.89\\\scriptsize{ ($\pm$0.07)}}} 
&\makecell{\shortstack{\rule{0in}{2ex}72.03\\\scriptsize{ ($\pm$0.01)}}}
& \makecell{\shortstack{\rule{0in}{2ex}72.24\\\scriptsize{ ($\pm$0.05)}}}
&\makecell{\shortstack{\rule{0in}{2ex}\textbf{72.27}\\\scriptsize{ ($\pm$0.06)}}} 
&\makecell{\shortstack{\rule{0in}{2ex}72.11\\\scriptsize{ ($\pm$0.07)}}} 
&\makecell{\shortstack{\rule{0in}{2ex}72.09\\\scriptsize{ ($\pm$0.06)}}} 
\\

\textbf{Top-5}
& 91.42 &89.07 & \makecell{\shortstack{\rule{0in}{2ex}89.99\\\scriptsize{ ($\pm$0.05)}}}
& \makecell{\shortstack{\rule{0in}{2ex}90.46\\\scriptsize{ ($\pm$0.15)}}}     
&\makecell{\shortstack{\rule{0in}{2ex}90.21\\\scriptsize{ ($\pm$0.03)}}}
&    \makecell{\shortstack{\rule{0in}{2ex}90.28\\\scriptsize{ ($\pm$0.04)}}}
&    \makecell{\shortstack{\rule{0in}{2ex}90.59\\\scriptsize{ ($\pm$0.07)}}} 
&\makecell{\shortstack{\rule{0in}{2ex}90.85\\\scriptsize{ ($\pm$0.04)}}}
&\makecell{\shortstack{\rule{0in}{2ex}90.86\\\scriptsize{ ($\pm$0.08)}}}
&\makecell{\shortstack{\rule{0in}{2ex}\textbf{90.93}\\\scriptsize{ (\textbf{$\pm$0.03})}}}     
&\makecell{\shortstack{\rule{0in}{2ex}\textbf{90.95}\\\scriptsize{ (\textbf{$\pm$0.03})}}}   
&\makecell{\shortstack{\rule{0in}{2ex}\textbf{90.95}\\\scriptsize{ (\textbf{$\pm$0.03})}}}   
\\
\hline

		\end{tabular}
	}
	\caption{Top-1 and Top-5 accuracies (\%) and standard deviations of compression methods on ImageNet validation set. The teacher is ResNet34 and the student is ResNet18. 
	}
	\label{tab:imagenet_tab}%
\end{table*}%
\begin{table*}[h]
	\centering
	\begin{tabular}{|c|c|c|c|c|c|c|c|}
		\hline
		\bfseries\makecell{Teacher Network\\(Accuracy, \%)}
		& \makecell{WRN-40-2\\(76.35)}
		& \makecell{WRN-40-2\\(76.35)}
		& \makecell{ResNet56\\(73.23)}
		& \makecell{ResNet110\\(74.36)} 
		& \makecell{ResNet110\\(74.36)}
		& \makecell{ResNet32x4\\(79.54)}
		& \makecell{VGG13\\(74.93)}\\
		\bfseries\makecell{Student Network\\(Accuracy, \%)} 
		& \makecell{WRN-16-2\\(73.32)} 
		& \makecell{WRN-40-1\\(71.83)} 
		& \makecell{ResNet20\\(69.28)} 
		& \makecell{ResNet20\\(69.28)} 
		& \makecell{ResNet32\\(71.50)} 
		& \makecell{ResNet8x4\\(72.83)} 
		& \makecell{VGG8\\(70.92)} \\
		\hhline{|========|}
		\textbf{Method} & \multicolumn{7}{c|}{\textbf{Accuracy (std.)}} \\
		\Xhline{3\arrayrulewidth}
KD \cite{KD}&75.78\scriptsize{ ($\pm$0.09)}    &74.62\scriptsize{ ($\pm$0.19)}    &71.90\scriptsize{ ($\pm$0.43)}    &72.14\scriptsize{ ($\pm$0.20)}    &74.24\scriptsize{ ($\pm$0.09)}    &73.49\scriptsize{ ($\pm$0.29)}    &73.81\scriptsize{ ($\pm$0.12)} \\
FitNet \cite{FitNet}&73.72\scriptsize{ ($\pm$0.25)}    &72.48\scriptsize{ ($\pm$0.07)}    &69.09\scriptsize{ ($\pm$0.24)}    &69.41\scriptsize{ ($\pm$0.16)}    &70.73\scriptsize{ ($\pm$0.12)}    &73.24\scriptsize{ ($\pm$0.11)}    &71.35\scriptsize{ ($\pm$0.09)} \\
AT \cite{AT}&74.23\scriptsize{ ($\pm$0.17)}    &73.42\scriptsize{ ($\pm$0.02)}    &70.90\scriptsize{ ($\pm$0.03)}    &70.96\scriptsize{ ($\pm$0.11)}    &72.80\scriptsize{ ($\pm$0.25)}    &73.86\scriptsize{ ($\pm$0.06)}    &71.89\scriptsize{ ($\pm$0.26)}         \\
RKD \cite{RKD}&73.97\scriptsize{ ($\pm$0.32)}    &72.70\scriptsize{ ($\pm$0.18)}    &70.92\scriptsize{ ($\pm$0.25)}    &70.54\scriptsize{ ($\pm$0.31)}    &72.65\scriptsize{ ($\pm$0.11)}    &71.75\scriptsize{ ($\pm$0.18)}    &71.44\scriptsize{ ($\pm$0.07)} \\
FT \cite{FT}&73.69\scriptsize{ ($\pm$0.26)}    &72.14\scriptsize{ ($\pm$0.03)}    &70.77\scriptsize{ ($\pm$0.29)}    &70.91\scriptsize{ ($\pm$0.13)}    &72.36\scriptsize{ ($\pm$0.47)}    &73.53\scriptsize{ ($\pm$0.07)}    &71.29\scriptsize{ ($\pm$0.29)}\\
CC \cite{CC}&73.38\scriptsize{ ($\pm$0.18)}    &72.05\scriptsize{ ($\pm$0.41)}    &70.20\scriptsize{ ($\pm$0.31)}    &69.76\scriptsize{ ($\pm$0.17)}    &71.58\scriptsize{ ($\pm$0.02)}    &72.65\scriptsize{ ($\pm$0.16)}    &70.52\scriptsize{ ($\pm$0.10)}         \\
OFD \cite{overhaul}&75.32\scriptsize{ ($\pm$0.32)}    &74.21\scriptsize{ ($\pm$0.15)}    &71.48\scriptsize{ ($\pm$0.13)}    &71.59\scriptsize{ ($\pm$0.13)}    &73.25\scriptsize{ ($\pm$0.20)}    &74.35\scriptsize{ ($\pm$0.08)}    &71.13\scriptsize{ ($\pm$0.10)}\\
CRD \cite{CRD}&76.08\scriptsize{ ($\pm$0.09)}    &74.68\scriptsize{ ($\pm$0.22)}    &72.39\scriptsize{ ($\pm$0.30)}    &72.28\scriptsize{ ($\pm$0.24)}    &74.19\scriptsize{ ($\pm$0.10)}    &75.61\scriptsize{ ($\pm$0.07)}    &73.80\scriptsize{ ($\pm$0.16)}\\
CRD+KD \cite{CRD}&76.08\scriptsize{ ($\pm$0.12)}    &75.50\scriptsize{ ($\pm$0.14)}    &72.67\scriptsize{ ($\pm$0.11)}    &72.66\scriptsize{ ($\pm$0.09)}    &74.72\scriptsize{ ($\pm$0.15)}    &75.49\scriptsize{ ($\pm$0.10)}    &74.37\scriptsize{ ($\pm$0.07)}\\
ReviewKD \cite{reviewkd}&75.75\scriptsize{ ($\pm$0.07)}    &74.90\scriptsize{ ($\pm$0.15)}    &71.93\scriptsize{ ($\pm$0.09)}    &72.09\scriptsize{ ($\pm$0.12)}    &73.68\scriptsize{ ($\pm$0.10)}    &74.18\scriptsize{ ($\pm$0.22)}    &72.01\scriptsize{ ($\pm$0.01)}\\

\Xhline{3\arrayrulewidth}
\textbf{DCKD Net1}  &\textbf{76.85}\scriptsize{ ($\pm$0.07)}    &\textbf{75.89}\scriptsize{ ($\pm$0.10)}    &\textbf{73.09}\scriptsize{ ($\pm$0.15)}    &\textbf{73.07}\scriptsize{ ($\pm$0.07)}    &\textbf{75.33}\scriptsize{ ($\pm$0.05)} &\textbf{76.00}\scriptsize{ ($\pm$0.16)} &\textbf{74.90}\scriptsize{ ($\pm$0.19)} \\
\textbf{DCKD Net2}  &\textbf{76.62}\scriptsize{ ($\pm$0.09)}   
&\textbf{75.65}\scriptsize{ ($\pm$0.02)}    
&\textbf{72.81}\scriptsize{ ($\pm$0.11)}    
&\textbf{72.88}\scriptsize{ ($\pm$0.09)}    
&\textbf{75.10}\scriptsize{ ($\pm$0.06)} 
&\textbf{75.84}\scriptsize{ ($\pm$0.06)} 
&\textbf{74.76}\scriptsize{ ($\pm$0.24)} \\
\textbf{DCKD Net3}  &\textbf{76.52}\scriptsize{ ($\pm$0.10)}    &\textbf{75.56}\scriptsize{ ($\pm$0.09)}    &72.61\scriptsize{ ($\pm$0.18)}    &\textbf{72.70}\scriptsize{ ($\pm$0.04)}    &\textbf{74.87}\scriptsize{ ($\pm$0.13)} &75.60\scriptsize{ ($\pm$0.19)} &\textbf{74.58}\scriptsize{ ($\pm$0.18)} \\
\hline		

	\end{tabular}%
	\caption{Top-1 accuracies (\%) and standard deviations of various student models on CIFAR100 validation set that have architectures that are similar to their teacher models. 
		The best performances are highlighted in bold.
	}
	\label{tab:cifar1}%
\end{table*}%

{\bf Reverse Kullback-Leibler Divergence} 
We discussed the optimal way to collect knowledge to encapsulate the necessary information about the relationships between classes. Now, we expand this idea further and obtain an adequate loss for the optimization.

The most frequently employed criterion for comparing probability distributions $\mathbf{u}$ and $\mathbf{v}$ is the Kullback-Leibler divergence (KLD) loss $\mathcal{L}_{KLD}$.
\begin{equation}\label{KLD}
	\mathcal{L}_{KLD}(\mathbf{u},\mathbf{v})=\mathcal{L}_{CE}(\mathbf{u},\mathbf{v})-H(\mathbf{u}),
\end{equation}
%
where $H(\mathbf{u})$ is the entropy of distribution $\mathbf{u}$.
\begin{equation}
	H(\mathbf{u})=-\sum_{c}\mathbf{u}_c\log \mathbf{u}_c.
\end{equation} 
When $\mathbf{u}$ is fixed, optimizing $\mathcal{L}_{CE}$ with distribution $\mathbf{v}$ is equivalent to optimizing $\mathcal{L}_{KLD}$. 
Note that $\mathcal{L}_{KLD}$ is asymmetric, \eg, $\mathcal{L}_{KLD}(\mathbf{u},\mathbf{v})\neq \mathcal{L}_{KLD}(\mathbf{v},\mathbf{u})$.

There are three ways to optimize $\mathcal{L}_{KLD}(\mathbf{u},\mathbf{v})$. First, $\mathbf{u}$ and $\mathbf{v}$ are simultaneously optimized with losses calculated in a bidirectional manner that makes $\mathbf{v}$ closer to $\mathbf{u}$ and $\mathbf{u}$ closer to $\mathbf{v}$. Second, $\mathbf{u}$ is fixed and only $\mathbf{v}$ is optimized. This method is referred to as a forward method and is applied in most optimization tasks using $\mathcal{L}_{KLD}(\mathbf{u},\mathbf{v})$. There is a reverse method of fixing $\mathbf{v}$ and optimizing only $\mathbf{u}$. We choose the reverse method to optimize our collection loss \eqref{backKLD} and will discuss more details. 

In \eqref{KLD}, to minimize $\mathcal{L}_{KLD}(\mathbf{u},\mathbf{v})$, there is a way to minimize $\mathcal{L}_{CE}(\mathbf{u},\mathbf{v})$, but also there is a way to increase the entropy $H(\mathbf{u})$.
$H(\mathbf{u})$ cannot be increased in the forward method because $\mathbf{u}$ is a fixed distribution. However, $H(\mathbf{u})$ can be increased in the reverse method because $ \mathbf{u} $ is optimized and $ \mathbf{v} $ is fixed.
It can be said that as the entropy increases, $ \mathbf{u} $ is smoothed and becomes a distribution that differs significantly from the one-hot distribution.

In the aforementioned study \cite{entropy}, the authors discussed the effects of high entropy on a model's output. 
Their main idea is that the high entropy of the model's distribution renders it more generalized and robust to gradient vanishing problems.
A teacher with smoothed outputs tends to have rich information on correlations between classes. 
\cite{max-entropy} explained that increasing the entropy of the model's distribution reduces the confidence of the model and leads to a better generalization.
Since low entropy limits the space to find optimal parameters, the authors showed that max-entropy regularization boosts the model’s performance. 
The increase in entropy improved the model’s representation for the correlations.

Therefore, instead of updating $\mathbf{v}$ in $\mathcal{L}_{KLD}(\mathbf{u},\mathbf{v})$, we considered the reverse method to update $\mathbf{u}$ for our collection loss. Our experiments indicated that students perform better when using the reverse KL divergence instead of the standard forward KL divergence.


\begin{table*}
	\centering
	\begin{tabular}{|c|c|c|c|c|c|c|}
		\hline
		\bfseries\makecell{Teacher Network\\(Accuracy, \%)}
		& \makecell{VGG13\\(74.93)}
		& \makecell{ResNet50\\(79.68)}
		& \makecell{ResNet50\\(79.68)}
		& \makecell{ResNet32x4\\(79.54)} 
		& \makecell{ResNet32x4\\(79.54)}
		& \makecell{WRN-40-2    \\(76.35)} \\
		\bfseries\makecell{Student Network\\(Accuracy, \%)} 
		& \makecell{MobileNetV2\\(65.23)} 
		& \makecell{MobileNetV2\\(65.23)} 
		& \makecell{VGG8\\(70.92)} 
		& \makecell{ShuffleNetV1\\(71.34)} 
		& \makecell{ShuffleNetV2\\(73.89)} 
		& \makecell{ShuffleNetV1\\(71.34)} \\
		\hhline{|=======|}
		\textbf{Method} & \multicolumn{6}{c|}{\textbf{Accuracy (std.)}} \\
		\Xhline{3\arrayrulewidth}
KD \cite{KD}       
&69.31\scriptsize{ ($\pm$0.41)}    
&69.46\scriptsize{ ($\pm$0.28)}    
&74.26\scriptsize{ ($\pm$0.06)}    
&75.43\scriptsize{ ($\pm$0.42)}    
&75.80\scriptsize{ ($\pm$0.34)}    
&76.63\scriptsize{ ($\pm$0.29)} \\

FitNet \cite{FitNet}    
&63.41\scriptsize{ ($\pm$0.92)}    
&62.11\scriptsize{ ($\pm$0.25)}    
&69.46\scriptsize{ ($\pm$0.06)}    
&73.74\scriptsize{ ($\pm$0.30)}    
&74.15\scriptsize{ ($\pm$0.11)}    
&74.32\scriptsize{ ($\pm$0.19)} \\

AT \cite{AT}       
&64.34\scriptsize{ ($\pm$0.12)}    
&63.73\scriptsize{ ($\pm$0.30)}    
&72.44\scriptsize{ ($\pm$0.18)}    
&74.40\scriptsize{ ($\pm$0.18)}    
&74.28\scriptsize{ ($\pm$0.09)}    
&75.00\scriptsize{ ($\pm$0.07)} \\

RKD \cite{RKD}      
&65.55\scriptsize{ ($\pm$0.53)}    
&65.88\scriptsize{ ($\pm$0.35)}    
&71.69\scriptsize{ ($\pm$0.17)}    
&74.17\scriptsize{ ($\pm$0.20)}    
&75.16\scriptsize{ ($\pm$0.17)}    
&75.12\scriptsize{ ($\pm$0.09)} \\

FT \cite{FT}       
&65.70\scriptsize{ ($\pm$0.34)}    
&65.35\scriptsize{ ($\pm$0.35)}    
&71.78\scriptsize{ ($\pm$0.10)}    
&74.56\scriptsize{ ($\pm$0.09)}    
&74.98\scriptsize{ ($\pm$0.09)}    
&74.91\scriptsize{ ($\pm$0.31)} \\

CC \cite{CC}       
&64.75\scriptsize{ ($\pm$0.16)}    
&65.32\scriptsize{ ($\pm$0.12)}    
&70.37\scriptsize{ ($\pm$0.21)}    
&72.48\scriptsize{ ($\pm$0.35)}    
&73.81\scriptsize{ ($\pm$0.36)}    
&72.75\scriptsize{ ($\pm$0.34)} \\

OFD \cite{overhaul}      
&63.43\scriptsize{ ($\pm$0.74)}    
&63.32\scriptsize{ ($\pm$0.17)}    
&71.79\scriptsize{ ($\pm$0.36)}    
&75.56\scriptsize{ ($\pm$0.25)}    
&76.81\scriptsize{ ($\pm$0.17)}    
&75.94\scriptsize{ ($\pm$0.19)} \\

CRD \cite{CRD}      
&69.38\scriptsize{ ($\pm$0.45)}    
&69.22\scriptsize{ ($\pm$0.15)}    
&74.10\scriptsize{ ($\pm$0.02)}    
&76.01\scriptsize{ ($\pm$0.09)}    
&76.06\scriptsize{ ($\pm$0.03)}    
&76.66\scriptsize{ ($\pm$0.11)} \\

CRD+KD \cite{CRD}    
&70.57\scriptsize{ ($\pm$0.07)}    
&71.06\scriptsize{ ($\pm$0.09)}    
&74.60\scriptsize{ ($\pm$0.12)}    
&76.42\scriptsize{ ($\pm$0.29)}    
&77.18\scriptsize{ ($\pm$0.28)}    
&77.45\scriptsize{ ($\pm$0.38)} \\

ReviewKD \cite{reviewkd}    
&67.58\scriptsize{ ($\pm$0.40)}    
&66.62\scriptsize{ ($\pm$0.16)}    
&70.86\scriptsize{ ($\pm$0.24)}    
&75.85\scriptsize{ ($\pm$0.18)}    
&76.07\scriptsize{ ($\pm$0.27)}    
&76.94\scriptsize{ ($\pm$0.11)} \\

\Xhline{3\arrayrulewidth}
\textbf{DCKD Net1}     
&\textbf{71.02}\scriptsize{ ($\pm$0.08)}    
&\textbf{71.40}\scriptsize{ ($\pm$0.11)} 
&\textbf{75.05}\scriptsize{ ($\pm$0.03)} 
&\textbf{76.75}\scriptsize{ ($\pm$0.08)}    
&\textbf{77.60}\scriptsize{ ($\pm$0.03)}    
&\textbf{77.89}\scriptsize{ ($\pm$0.07)}\\
\textbf{DCKD Net2}     
&\textbf{70.78}\scriptsize{ ($\pm$0.12)}    
&70.97\scriptsize{ ($\pm$0.38)} 
&\textbf{74.75}\scriptsize{ ($\pm$0.10)} 
&\textbf{76.47}\scriptsize{ ($\pm$0.12)}    
&\textbf{77.49}\scriptsize{ ($\pm$0.07)}    
&77.41\scriptsize{ ($\pm$0.17)}\\

\textbf{DCKD Net3}     
&\textbf{70.67}\scriptsize{ ($\pm$0.08)}    
&70.73\scriptsize{ ($\pm$0.33)} 
&74.56\scriptsize{ ($\pm$0.06)} 
&76.34\scriptsize{ ($\pm$0.02)}    
&\textbf{77.26}\scriptsize{ ($\pm$0.11)}   
&77.25\scriptsize{ ($\pm$0.16)}\\

		\hline
	\end{tabular}%
	\caption{Top-1 accuracies (\%) and standard deviations of compressed student models on CIFAR-100 validation set that have architectures that are different from their teacher models.
		The best performances are highlighted in bold.
	}
	\label{tab:cifar2}%
\end{table*}%

\section{Experiments}

{\bf Datasets}
We use four datasets for the image classification task: ImageNet \cite{imagenet}, CIFAR-100, CIFAR-10 \cite{cifar} and Fashion-MNIST \cite{FMNIST}. 
ImageNet has 1K classes, each with 1.2M training images and 50K validation images.
Each of CIFAR-10 and CIFAR-100 contains 50K train images and 10K validation images. 
Fashion-MNIST consists of 60K train and 10K validation grayscale images of 10 classes.

{\bf Setup} 
For all experiments, we use the stochastic gradient descent (SGD) with cosine learning rate scheduler \cite{coslr}, where $T_0$ is 30 and $T_{mult}$ is $2$.
The number of training epochs for ImageNet, CIFAR-100, CIFAR-10 and Fashion-MNIST are 210, 930, 210 and 450, respectively. 
All experiments are conducted three times; their average accuracies and standard deviations are summarized in tables.
In supplementary materials, more detailed results are reported for all experiments.

For hyperparameters, we empirically use $\beta_{CE}$ as $1$, $\beta_{KD}$ as $1$ and $\beta_{Col}$ as $0.2$ for ImageNet. Otherwise, we use $\beta_{Col}$ as $0.5$.
The temperature of the knowledge distillation loss $T_{KD}$ is set to $4$, and the temperature of the collection loss $T_{KLD}$ is set to $2$.
When training DCKD, we use three untrained student models that are randomly initialized.

\subsection{Model Compression}
{\bf Compression methods} We compare seven knowledge distillation methods with our DCKD for model compression: Knowledge Distillation (KD) \cite{KD},
FitNets: Hints for thin deep nets (FitNet) \cite{FitNet},
Attention Transfer (AT) \cite{AT},
Relational Knowledge Distillation (RKD) \cite{RKD},
network compression via Factor Transfer (FT) \cite{FT}, Correlation Congruence (CC) \cite{CC}, a comprehensive Overhaul of Feature Distillation (OFD) \cite{overhaul}, Contrastive Representation Distillation (CRD) \cite{CRD} and distilling knowledge via knowledge review (ReviewKD) \cite{reviewkd}.
We follow the implementation details utilized in \cite{reviewkd, overhaul, CRD} for a fair comparison.

{\bf Results on ImageNet} In experiments on ImageNet, we follow the standard training parameters of ImageNet in PyTorch. 
In \cref{tab:imagenet_tab}, we compare KD, AT, CC, OFD, CRD and ReviewKD with our DCKD. 
We use pretrained ResNet34 as the teacher model and ResNet18 as the student model.
The performance of DCKD is superior to that of other compared methods, and these results indicate that DCKD works well on large datasets.

{\bf Results on CIFAR-100}
We performed experiments with various network architectures on CIFAR-100.
The initial learning rate for MobileNet \cite{mobnet} and ShuffleNet \cite{shunet} is set to $0.01$; otherwise, it is set to $0.05$.
To compare the distillation methods between similar network architectures, we employ them in ResNet \cite{resnet}, Wide ResNet \cite{wrn} and VGG \cite{VGG}. The average validation accuracies are summarized in \cref{tab:cifar1}.
Our DCKD results show excellent performance in all cases of the similar architectures. 

We focus on lighter models, such as MobileNet \cite{mobnet} and ShuffleNet \cite{shunet}, to compare them between different network architectures. The average validation accuracies are summarized in \cref{tab:cifar2}.
The results of model compression between different architectures show that our method has better performance than other compression methods.

\begin{table*}
	\begin{center}
		\resizebox{\textwidth}{!}{
			\renewcommand{\tabcolsep}{2.6mm}
			\begin{tabular}{|l|cc||ccccc?ccc|}
				\hline
				\multirow{2}{*}{\textbf{Dataset}}&\multirow{2}{*}{\textbf{Teacher}}&\multirow{2}{*}{\textbf{Student}}&	\multirow{2}{*}{KD \cite{KD}}	&\multirow{2}{*}{AT \cite{AT}}&\multirow{2}{*}{CC \cite{CC}}&
				\multirow{2}{*}{CRD	\cite{CRD}}&\multirow{2}{*}{CRD+KD \cite{CRD}}&\multicolumn{3}{c|}{\textbf{DCKD}}\\
				&&&&&&&&\textbf{Net1} &\textbf{Net2} &\textbf{Net3}\\
				\hline
				CIFAR-10&92.10&91.56
				&\makecell{\shortstack{\rule{0in}{2ex}93.75\\\scriptsize{ ($\pm$0.22)}}}
				&\makecell{\shortstack{\rule{0in}{2ex}92.86\\\scriptsize{ ($\pm$0.07)}}}
				&\makecell{\shortstack{\rule{0in}{2ex}93.74\\\scriptsize{ ($\pm$0.14)}}}
				&\makecell{\shortstack{\rule{0in}{2ex}90.35\\\scriptsize{ ($\pm$0.24)}}}
				&\makecell{\shortstack{\rule{0in}{2ex}92.96\\\scriptsize{ ($\pm$0.09)}}}
				&\makecell{\shortstack{\rule{0in}{2ex}\textbf{94.09}\\\scriptsize{ ($\pm$0.02)}}}   
				&\makecell{\shortstack{\rule{0in}{2ex}\textbf{93.91}\\\scriptsize{ ($\pm$0.06)}}}   
				&\makecell{\shortstack{\rule{0in}{2ex}\textbf{93.82}\\\scriptsize{ ($\pm$0.10)}}}   \\
				\hline
				Fashion-MNIST&94.08&93.22
				&\makecell{\shortstack{\rule{0in}{2ex}94.30\\\scriptsize{ ($\pm$0.05)}}}
				&\makecell{\shortstack{\rule{0in}{2ex}94.21\\\scriptsize{ ($\pm$0.11)}}}
				&\makecell{\shortstack{\rule{0in}{2ex}94.21\\\scriptsize{ ($\pm$0.14)}}}
				&\makecell{\shortstack{\rule{0in}{2ex}91.68\\\scriptsize{ ($\pm$0.14)}}}
				&\makecell{\shortstack{\rule{0in}{2ex}93.27\\\scriptsize{ ($\pm$0.09)}}}
				&\makecell{\shortstack{\rule{0in}{2ex}\textbf{94.59}\\\scriptsize{ ($\pm$0.06)}}}
				&\makecell{\shortstack{\rule{0in}{2ex}\textbf{94.47}\\\scriptsize{ ($\pm$0.08)}}}
				&\makecell{\shortstack{\rule{0in}{2ex}\textbf{94.34}\\\scriptsize{ ($\pm$0.06)}}} \\
				
				\hline
				
			\end{tabular}
		}
	\end{center}
	\caption{Top-1 validation accuracies (\%) and standard deviations of compression methods on small datasets. 
		The teacher is ResNet110 and the student is ResNet32.
	}
	\label{tab:small_data}
\end{table*}

{\bf Results on small datasets} 
We compare our performance in small datasets, including CIFAR-10 and Fashion-MNIST. The methods for comparison are KD, AT, CC and CRD. 
The averages of our experiment's results are summarized in \cref{tab:small_data}.
Our method consistently outperforms other methods throughout various datasets. It is notable that some methods
lose performance on smaller datasets, whereas AT and CC maintain comparatively higher performance.

\newcolumntype{?}{!{\vrule width 1pt}}
\begin{table}[h]
	\begin{center}
		\renewcommand{\tabcolsep}{1.7mm}
		\begin{tabular}{|c|ccc|}
			\hline
			\textbf{Method} & Net1 & Net2 &Net3\\
			\hline
			DML \cite{DML} & 74.60 \scriptsize{ ($\pm$0.20)} & 74.41 \scriptsize{ ($\pm$0.12)}& 74.32 \scriptsize{ ($\pm$0.09)}\\
			ONE \cite{ONE} & 74.19 \scriptsize{ ($\pm$0.18)}& 74.00 \scriptsize{ ($\pm$0.23)}& 73.76 \scriptsize{ ($\pm$0.25)}\\
			OKDDip \cite{OKDDIP} & 74.87 \scriptsize{ ($\pm$0.05)}& 74.58 \scriptsize{ ($\pm$0.09)}& 74.44 \scriptsize{ ($\pm$0.03)}\\
			KDCL \cite{KDCL} & 73.42 \scriptsize{ ($\pm$0.02)}& 73.22 \scriptsize{ ($\pm$0.06)}& 73.05 \scriptsize{ ($\pm$0.11)}\\
			\Xhline{3\arrayrulewidth}
			\textbf{DCKD}  & \textbf{75.33} \scriptsize{ ($\pm$0.05)}& \textbf{75.10} \scriptsize{ ($\pm$0.06)}& \textbf{74.87} \scriptsize{ ($\pm$0.13)}\\
			
			\hline
		\end{tabular}
	\end{center}
	\caption{Top-1 validation accuracies (\%) and standard deviations of the multistudent methods with the teacher except ONE.
	}
	\label{tab:multstu}
\end{table}

\subsection{Comparison with Multistudent Methods}

%
%

The architecture of DCKD is similar to mutual learning research because multiple student models share their knowledge with each other.
Therefore, we compare the following mutual learning methods with our method: Deep Mutual Learning (DML) \cite{DML}, knowledge distillation by On-the-fly Native Ensemble (ONE) \cite{ONE}, Online Knowledge Distillation with Diverse peers (OKDDip) \cite{OKDDIP} and online Knowledge Distillation via Collaborative Learning (KDCL) \cite{KDCL}.
We follow the implementation details utilized in \cite{OKDDIP, KDCL}.
Usually, these multistudent methods focus on the environment where a teacher is absent, but these methods are approaches that enable us to attach the teacher model, except ONE. 
Thus, we add the pretrained teacher model to DML and OKDDip for a fair comparison.
For KDCL, we used the pretrained teacher for a large network because KDCL is a method of training a large network and several small networks from scratch.

We use ResNet110 trained with CIFAR100 as the teacher model and three untrained ResNet32 as student models. 
We attach knowledge distillation loss to each student, as depicted in \cite{OKDDIP, DML}.
All methods have three student models, and the other hyperparameters are the same as those in the previous experiments.
The average accuracies and standard deviations of the students are summarized in \cref{tab:multstu}.
Our method outperforms the multistudent methods even when they are guided by the teacher network.

\subsection{Ablative Study}

In this section, we discuss the effects of our key components: direction of KL divergence, collecting methods and number of students.
The experiments are conducted using ResNet110 trained with CIFAR-100 as the teacher and using three ResNet32 as students. 
The baseline model is trained using the reverse KL divergence and the logit max collection method.
Other hyperparameters are identical to that of the experiments in \cref{tab:cifar1}.
All experiments are conducted three times; their average validation accuracies and standard deviations are summarized in tables.
\begin{figure*}
	\centering
	\begin{subfigure}{1\columnwidth}
		\centering
		\includegraphics[width=0.8\linewidth]{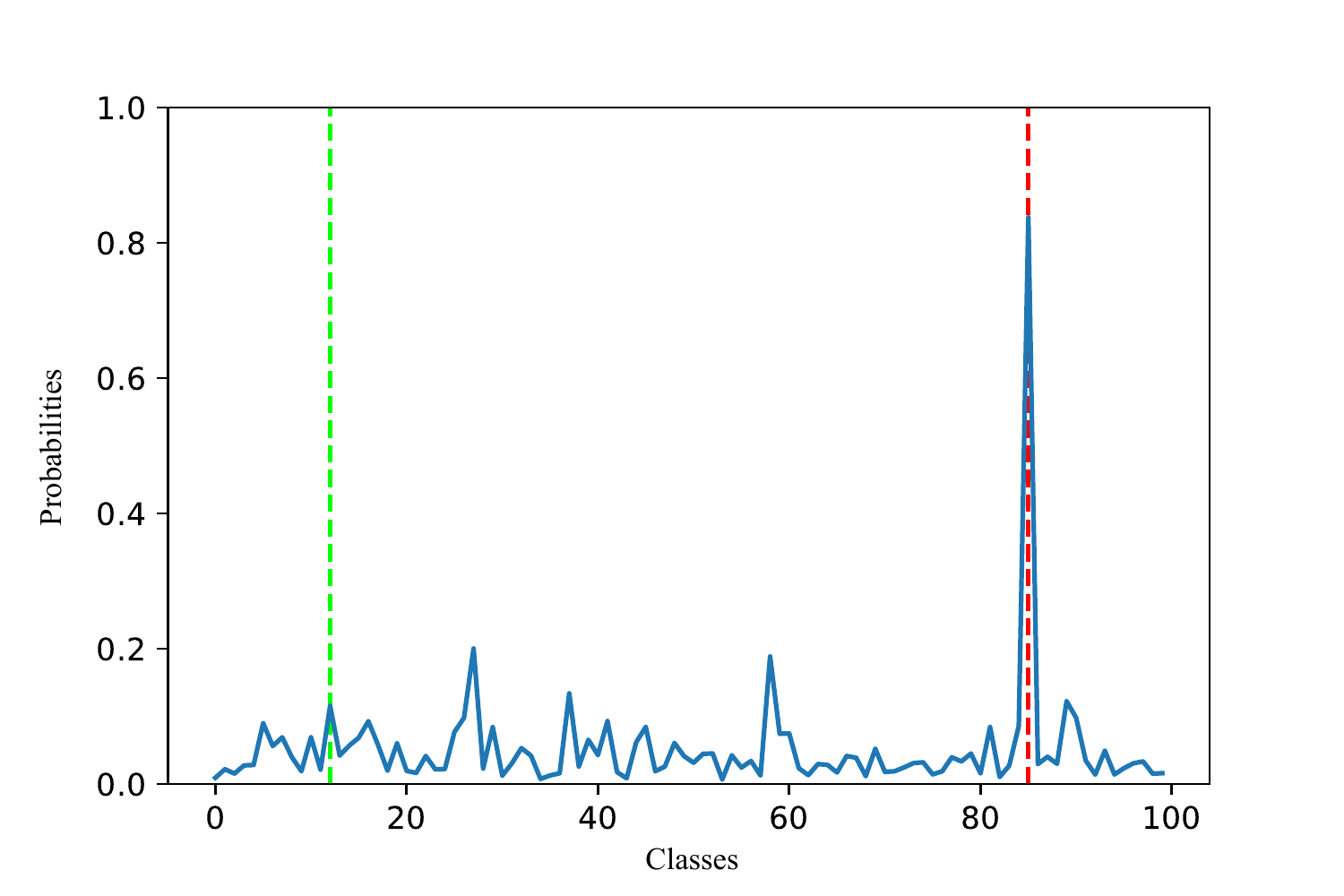}
		\caption{Accumulation of teacher outputs}
		\label{fig:Mentor}
	\end{subfigure}%
	\begin{subfigure}{1\columnwidth}
		\centering
		\includegraphics[width=0.8\linewidth]{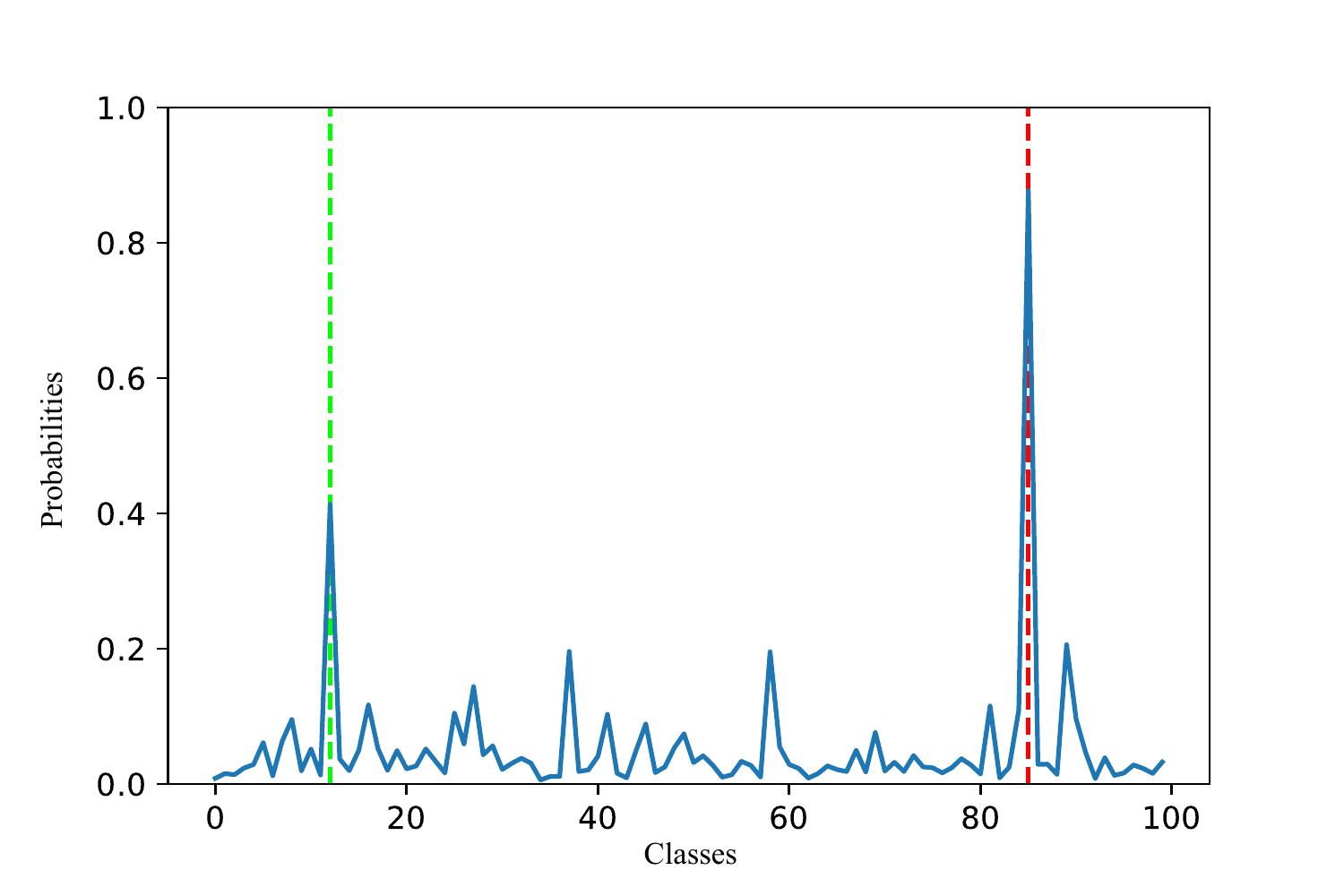}
		\caption{Accumulation of student outputs}
		\label{fig:Mentee}
	\end{subfigure}
	\caption{Accumulation of outputs of tank class images. The green line represents the bridge class (12th) and the red line represents the tank class (85th).}
	
\end{figure*}

\begin{table}[h]
	\begin{center}
		
		\begin{tabular}{|l|ccc|}
			\hline
			\textbf{Method}	&Net1&	Net2	&Net3	\\
			\hline
			
			DCKD & \makecell{\shortstack{\rule{0in}{2ex}75.33\\ \scriptsize{ ($\pm$0.05)}}}  
			& \makecell{\shortstack{\rule{0in}{2ex}75.10\\ \scriptsize{ ($\pm$0.06)}}} 
			& \makecell{\shortstack{\rule{0in}{2ex}74.87\\ \scriptsize{ ($\pm$0.13)}}} \\
			DCKD w/o Rev. KL & \makecell{\shortstack{\rule{0in}{2ex}75.01\\ \scriptsize{ ($\pm$0.06)}}}  
			& \makecell{\shortstack{\rule{0in}{2ex}74.72\\ \scriptsize{ ($\pm$0.16)}}} 
			& \makecell{\shortstack{\rule{0in}{2ex}74.57\\ \scriptsize{ ($\pm$0.06)}}} \\
			DCKD w/o Max Col. & \makecell{\shortstack{\rule{0in}{2ex}75.05\\ \scriptsize{ ($\pm$0.03)}}}  
			& \makecell{\shortstack{\rule{0in}{2ex}74.98\\ \scriptsize{ ($\pm$0.03)}}} 
			& \makecell{\shortstack{\rule{0in}{2ex}74.81\\ \scriptsize{ ($\pm$0.03)}}} \\
			DCKD w/o Rev. KL, Max Col. & \makecell{\shortstack{\rule{0in}{2ex}74.89\\ \scriptsize{ ($\pm$0.02)}}}  
			& \makecell{\shortstack{\rule{0in}{2ex}74.83\\ \scriptsize{ ($\pm$0.08)}}} 
			& \makecell{\shortstack{\rule{0in}{2ex}74.69\\ \scriptsize{ ($\pm$0.10)}}} \\
			\hline
		\end{tabular}
	\end{center}
	\caption{
		Rev. KL: reverse KL divergence.
		Max Col.: logit max collection method.}
	\label{tab:ablative}
\end{table}

{\bf Direction of KL divergence and collecting methods}
We performed ablation experiments with the following three ways to show the benefits of DCKD's components (\cref{tab:ablative}). 
For each experiment, we used three untrained student models that were randomly initialized.
(1) DCKD without Rev. KL: An experiment with standard forward KL divergence and logit max collection method performs less than the baseline model. 
The differences in accuracies for each student are 0.30$\sim$0.38\%, and it can be seen that the direction of KL divergence affects performance improvement.
(2) DCKD without Max Col.: An experiment with reverse KL divergence and the average collection method \eqref{avg} was performed to show the effect of the max collection method \eqref{log_max} on increasing the correlation information.
\eqref{log_max} and \eqref{p_max} have the same purpose, so the experimental results will be omitted.
(3) DCKD without Rev. KL, Max Col.: The average accuracies of student models using forward KL divergence and the average collection method are 0.18$\sim$0.44\% lower than the baseline accuracies.
\begin{table}[h]
	\begin{center}

		\begin{tabular}{|c|ccccc|}
			\hline
			\makecell{\shortstack{\textbf{\# of }\\\textbf{students}}}	&Net1&	Net2	&Net3	&Net4 &Net5 \\
			\hline
			

N=2 & \makecell{\shortstack{\rule{0in}{2ex}75.17\\ \scriptsize{ ($\pm$0.03)}}}  
& \makecell{\shortstack{\rule{0in}{2ex}74.85\\ \scriptsize{ ($\pm$0.05)}}} & & & \\
N=3 & \makecell{\shortstack{\rule{0in}{2ex}75.33\\ \scriptsize{ ($\pm$0.05)}}}  
& \makecell{\shortstack{\rule{0in}{2ex}75.10\\ \scriptsize{ ($\pm$0.06)}}} 
& \makecell{\shortstack{\rule{0in}{2ex}74.87\\ \scriptsize{ ($\pm$0.13)}}} 
& & \\
N=4 & \makecell{\shortstack{\rule{0in}{2ex}75.06\\ \scriptsize{ ($\pm$0.03)}}}  
& \makecell{\shortstack{\rule{0in}{2ex}74.89\\ \scriptsize{ ($\pm$0.08)}}} 
& \makecell{\shortstack{\rule{0in}{2ex}74.81\\ \scriptsize{ ($\pm$0.07)}}} 
& \makecell{\shortstack{\rule{0in}{2ex}74.72\\ \scriptsize{ ($\pm$0.00)}}} 
& \\
N=5 & \makecell{\shortstack{\rule{0in}{2ex}75.05\\ \scriptsize{ ($\pm$0.03)}}}  
& \makecell{\shortstack{\rule{0in}{2ex}74.96\\ \scriptsize{ ($\pm$0.07)}}} 
& \makecell{\shortstack{\rule{0in}{2ex}74.92\\ \scriptsize{ ($\pm$0.08)}}} 
& \makecell{\shortstack{\rule{0in}{2ex}74.82\\ \scriptsize{ ($\pm$0.05)}}} 
& \makecell{\shortstack{\rule{0in}{2ex}74.64\\ \scriptsize{ ($\pm$0.10)}}} \\
			\hline
		\end{tabular}
	\end{center}
	\caption{Top-1 validation accuracies (\%) and standard deviations of DCKD per the number of students. N is the number of students.}
	\label{tab:numstu}
\end{table}

{\bf Number of students} 
We expected that as the number of students increased, the collective knowledge would become more accumulated and enriched. However, in the case of using more than four students, it seems that the students mimic each other and generate similar collective knowledge, which yields similar accuracy.
We believe performance gains can be expected if each student has different architecture or if students have been pretrained to be sufficiently different from each other.

\subsection{Discussion}

\begin{figure}
	\centering
	\begin{subfigure}{0.5\columnwidth}
		\centering
		\includegraphics[width=0.8\linewidth]{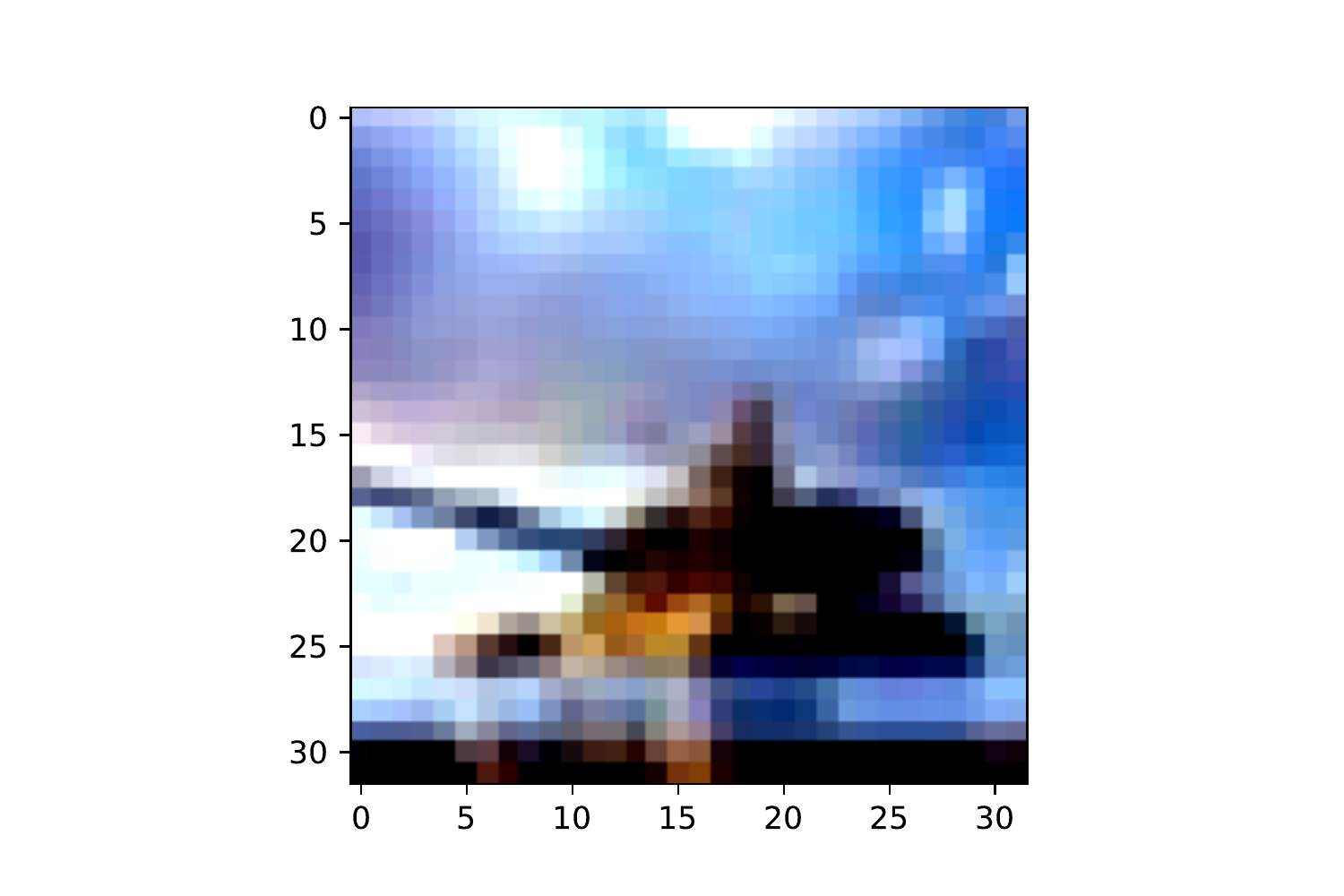}
		\caption{Class: Tank}
		\label{fig:tank}
	\end{subfigure}%
	\hspace*{-1.0cm}   
	\begin{subfigure}{0.5\columnwidth}
		\centering
		\includegraphics[width=0.8\linewidth]{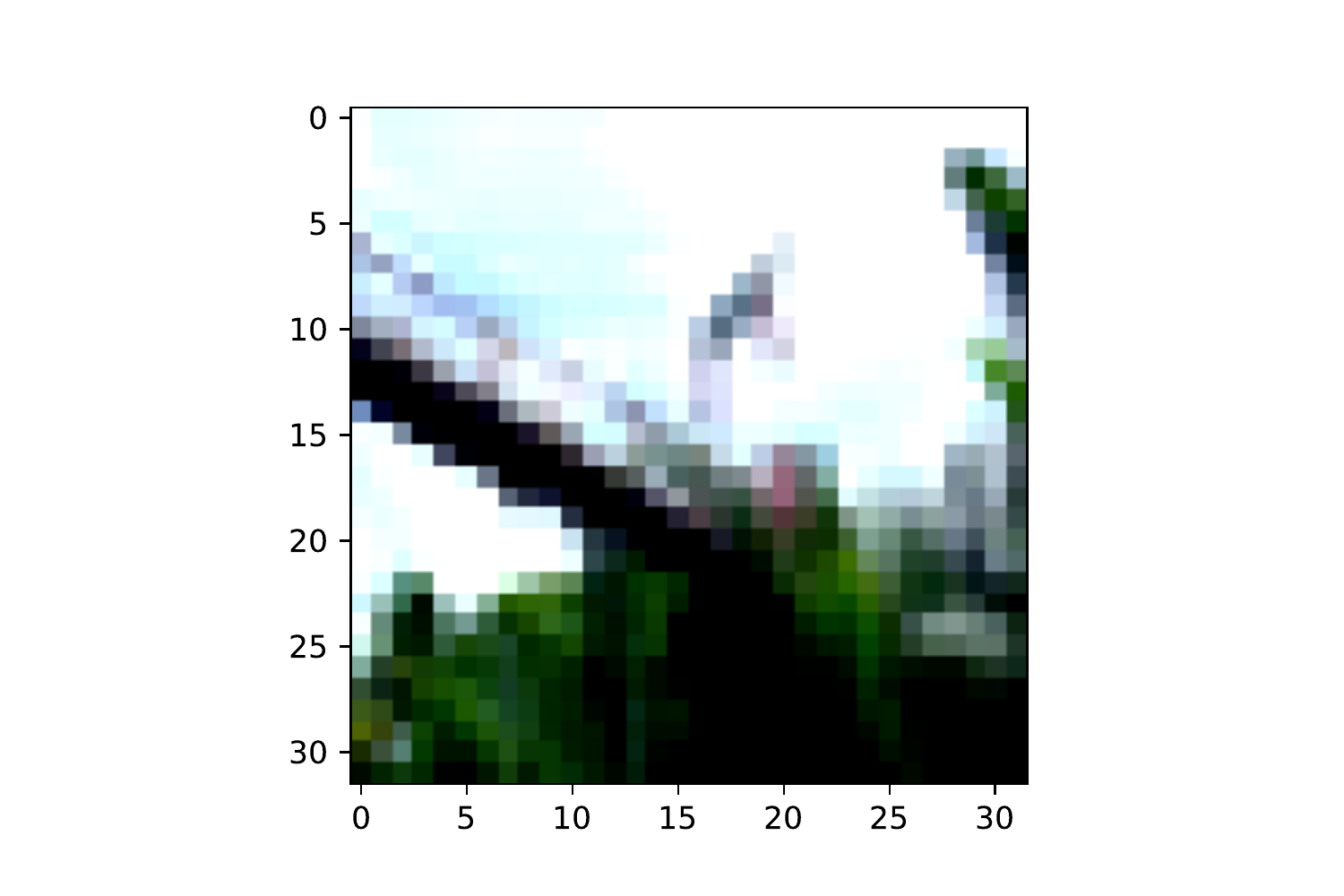}
		\caption{Class: Bridge}
		\label{fig:bridge}
	\end{subfigure}%
	\caption{Sample images \cite{cifar} from tank and bridge classes. }\label{fig:1}
\end{figure}

{\bf Visualization of correlation}
We visualize the aggregation of the teacher's outputs and the student's outputs.
\cref{fig:Mentor} and \cref{fig:Mentee} are obtained by applying the maximum function to the outputs $\bf{p}$ of each of the teacher (ResNet110) and one of the DCKD's students (ResNet32) for all images of the tank class in CIFAR-100. 
To achieve a clearer distribution, we use $\bf{p}=softmax(\bf{y}/T)$, where $T$ is $4$.
Both distributions show peaks at their ground truth class and at other classes, which may correlate with the tank images.
Although the teacher's distribution and student's distribution are similar in most classes, we observe another peak in the student's distribution that does not appear in the teacher's distribution.
\cref{fig:tank} shows the input data of the tank class that caused the peak for the bridge class in \cref{fig:Mentee}. The right side of \cref{fig:bridge} shows the sample input data of the bridge class, and since \cref{fig:tank} and \cref{fig:bridge} appear similar, the student has a peak in the bridge class.
The teacher model is well trained because it does not mistake tanks and bridges, but the well-trained teacher limits students from learning the correlation that reveals that tanks and bridges have similar features.
These correlations, which the teacher is trained to disregard, can be important for students who seek to obtain as much information as possible.

\begin{table*}[h]
	\centering
	\renewcommand{\tabcolsep}{1.5mm}
				\resizebox{\textwidth}{!}{
	\begin{tabular}{|c|c|c|c|c|c|c|c|}
		\hline
		\bfseries\makecell{Teacher (Accuracy, \%)\\DCKD Net1\\DCKD Net2\\DCKD Net3} 
		& \makecell{WRN-16-2\\(76.85)\\(76.62)\\(76.52)} 
		& \makecell{ResNet20\\(73.09)\\(72.81)\\(72.61)} 
		& \makecell{ResNet32\\(75.33)\\(75.10)\\(74.87)} 
		& \makecell{VGG8\\(74.90)\\(74.76)\\(74.58)}
		& \makecell{MobileNetV2\\(71.02)\\(70.78)\\(70.67)}
		& \makecell{ShuffleNetV1\\(77.89)\\(77.41)\\(77.25)}
		& \makecell{ShuffleNetV2\\(77.60)\\(77.49)\\(77.26)} \\
		\bfseries\makecell{Student Network\\(Accuracy, \%)} 
		& \makecell{WRN-16-2\\(73.32)} 
		& \makecell{ResNet20\\(69.28)} 
		& \makecell{ResNet32\\(71.50)} 
		& \makecell{VGG8\\(70.92)}
		& \makecell{MobileNetV2\\(65.23)}
		& \makecell{ShuffleNetV1\\(71.34)}
		& \makecell{ShuffleNetV2\\(73.89)} \\
		
		\hhline{|========|}
		\textbf{Method} & \multicolumn{7}{c|}{\textbf{Accuracy (std.)}} \\
		\Xhline{3\arrayrulewidth}
		
		\textbf{eDCKD Net1} 
		&\textbf{77.19}\scriptsize{ ($\pm$0.08)}	
		&72.70\scriptsize{ ($\pm$0.03)}	
		&\textbf{75.54}\scriptsize{ ($\pm$0.03)}  
		&\textbf{75.26}\scriptsize{ ($\pm$0.23)} 
		&\textbf{71.86}\scriptsize{ ($\pm$0.16)} 
		&77.51\scriptsize{ ($\pm$0.04)} 
		&77.37\scriptsize{ ($\pm$0.06)}\\
		
		\textbf{eDCKD Net2} 
		&\textbf{77.05}\scriptsize{ ($\pm$0.01)}	
		&72.59\scriptsize{ ($\pm$0.04)}	
		&75.33\scriptsize{ ($\pm$0.11)}  
		&\textbf{75.09}\scriptsize{ ($\pm$0.09)} 
		&\textbf{71.57}\scriptsize{ ($\pm$0.33)} 
		&77.28\scriptsize{ ($\pm$0.13)} 
		&77.11\scriptsize{ ($\pm$0.16)}\\
		
		\textbf{eDCKD Net3} 
		&\textbf{76.90}\scriptsize{ ($\pm$0.11)}	
		&72.43\scriptsize{ ($\pm$0.10)}	
		&75.24\scriptsize{ ($\pm$0.15)}  
		&\textbf{74.99}\scriptsize{ ($\pm$0.09)} 
		&\textbf{71.50}\scriptsize{ ($\pm$0.28)} 
		&77.00\scriptsize{ ($\pm$0.06)} 
		&76.93\scriptsize{ ($\pm$0.08)}\\
		\hline
		
		\textbf{Ensembled Student} 
		&\textbf{77.05}\scriptsize{ ($\pm$0.13)}	
		&72.61\scriptsize{ ($\pm$0.07)}	
		&75.28\scriptsize{ ($\pm$0.25)}  
		&\textbf{75.27}\scriptsize{ ($\pm$0.12)} 
		&\textbf{71.08}\scriptsize{ ($\pm$0.24)} 
		&77.87\scriptsize{ ($\pm$0.15)} 
		&77.26\scriptsize{ ($\pm$0.22)}\\
		
		\hline
	\end{tabular}%
	}
	\caption{Top-1 validation accuracies (\%) and standard deviations of various student models on CIFAR100.
		The teacher models are DCKD models in \cref{tab:cifar1,tab:cifar2}.
		The performances of eDCKD and ensembled student improved over DCKD are highlighted in bold.
	}
	\label{tab:edckd}%
\end{table*}%

{\bf Knowledge accumulation}
Our motivation is to capture the correlation information between classes that the teacher may overlook. 
We investigate the aggregation of knowledge that was collected from the teacher, and DCKD's students.
In previous knowledge distillation studies, entropy was employed as a standard metric to analyze the correlations between labels.
However, since the entropy is a measure of the degree of clustering in a distribution, it is not an ideal metric to evaluate correlations between classes.
For example, if there are two distributions of $\mathbf{p}_1=(0.6,0.05,\cdots,0.05)$ and  $\mathbf{p}_2=(0.6,0.4,\cdots,0)$, the entropy of $\mathbf{p}_1$ is higher than the entropy of $\mathbf{p}_2$, but $\mathbf{p}_1$ does not convey any correlation information such as that of $\mathbf{p}_2$.

\begin{figure}
	\centering
	\includegraphics[width=0.8\linewidth]{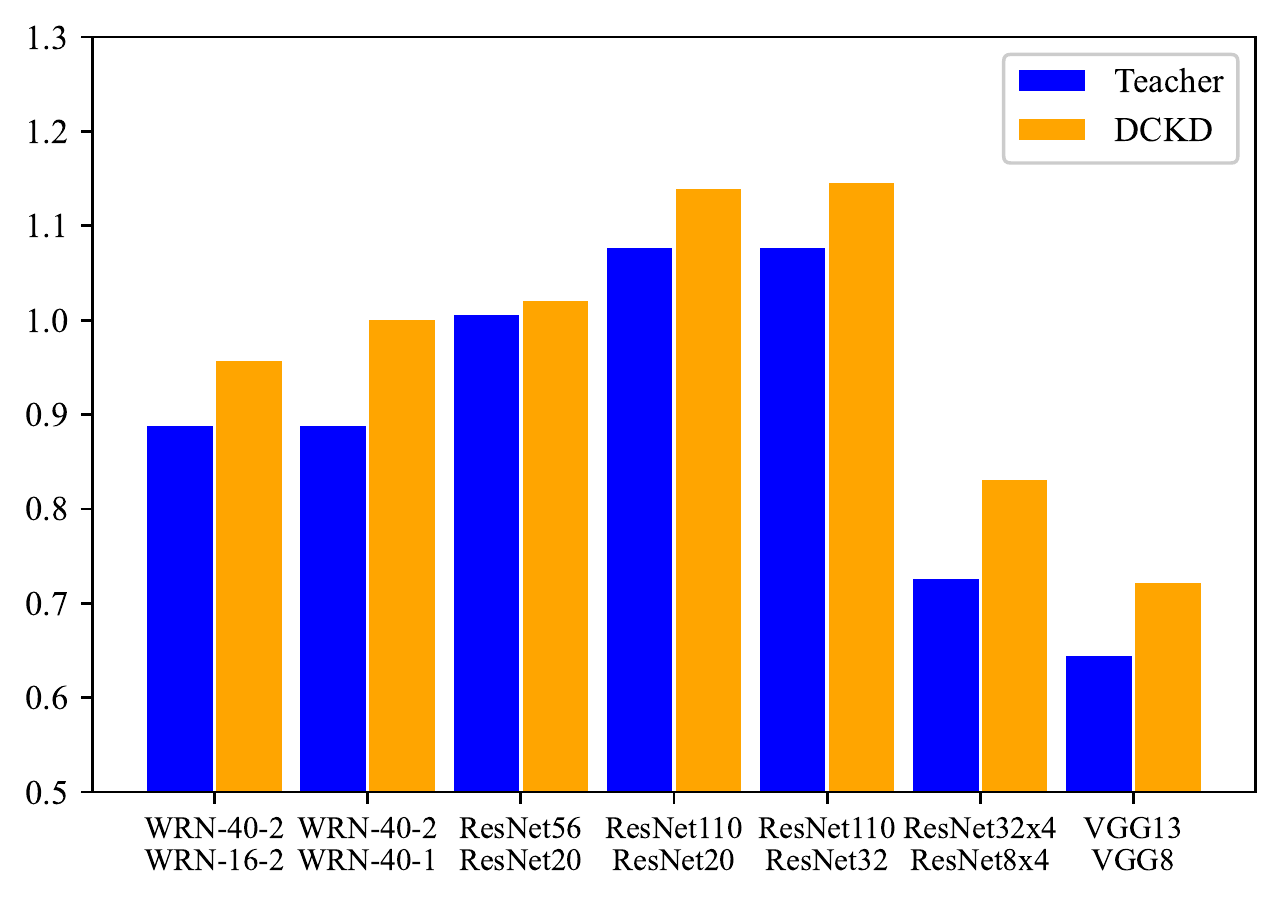}
	\caption{Average of the correlation number is calculated using the output distributions of teachers and students from \cref{tab:cifar1} softened by temperature $ 4 $, where $ threshold $ is $0.1$.}
	\label{fig:hist}
\end{figure}

Therefore, to better evaluate the correlation information, we propose an alternative metric that is referred to as the correlation number, which is defined as follows:
\begin{equation}
	K(\mathbf{p},threshold)=count(\{c|\mathbf{p}(c)>threshold\}),
\end{equation} 
where $\mathbf{p}$ is a distribution and $ c $ is a class index.
This metric counts the number of peaks in the distribution that are greater than $threshold$; therefore $K(\mathbf{p}_1,0.1)$ is 1 and $K(\mathbf{p}_2,0.1)$ is 2.
Thus, the metric $K$ counts the significant correlations between classes.
\cref{fig:hist} shows that average correlation numbers of DCKD are higher than those of teachers.
This suggests that DCKD's students have additional information compared with their teachers and the knowledge collected from the students significantly enrich the knowledge of DCKD.

{\bf Further distillation}
We showed that DCKD's students can be good teachers because they have a wealth of information on the correlations between classes. 
Therefore, we disseminate this accumulated knowledge from students with DCKD to the next generation of students via additional distillation, and we call this enhanced DCKD (eDCKD).
Instead of using DCKD's teacher, we utilize the students of DCKD as teachers for eDCKD's students.
We define that $\mathbf{p}_{T}$ of \eqref{KD_loss} is the ensembled output of the DCKD's students to utilize the accumulated knowledge considering the correlations between classes.
We train new student models of the eDCKD using \eqref{Ltot} with a newly defined $\mathbf{p}_{T}$. 

\cref{tab:edckd} shows that the performance of eDCKD is as good as that of DCKD although eDCKD's students were not trained with teachers of DCKD.
Moreover, some experiments show that eDCKD's students perform better than their teachers, DCKD's students.
Thus, DCKD's students have various correlational information, and the additional knowledge is helpful for teaching eDCKD's students.

In \cref{tab:edckd}, an ensembled student model is a single student model trained with the knowledge of three DCKD students.
Since we pursued a single compression model, we created an ensembled student model that combined the rich knowledge of multistudents into one.
The performances of ensembled students are improved over all the pretrained student models and some DCKD's students. 


\section{Conclusion}
We have proposed a series of new methods for knowledge distillation based on the notion that the teacher model may not have all of the knowledge that students need. 
To collect and transfer rich knowledge that the teacher does not provide, we utilized multistudent models and applied the reverse KL divergence to increase the entropy, which soften the output distributions. 
We applied our method to a variety of deep neural networks to observe a collection of knowledge that takes into account the relation between classes. 
Our simple, effective method for collecting and transferring knowledge has improved student models with a state-of-the-art performance across a variety of architectures and datasets.

{\small
\bibliographystyle{ieee_fullname}
\bibliography{egbib}
}

\end{document}


\title{Supplementary Materials}
	
	\maketitle

\section{Tables}
Here, we summarize all experimental results. We performed experiments on four datasets: ImageNet, CIFAR-100, CIFAR-10 and Fashion-MNIST. For all experiments, we use the stochastic gradient descent (SGD) with cosine annealing learning rate scheduler, where $T_0$ is 30 and $T_{mult}$ is $2$. All experiments are conducted three times.
For the hyperparameters of compared methods, we follow the implementation details utilized in CRD (Tian \etal, 2019), OFD (Heo \etal, 2019) and ReviewKD (Zhao \etal 2021).
Code will be released after publishing the paper due to company policies.

\subsection{ImageNet}
For all experiments, we use two GPUs for training with batch size of 512. The initial learning rate is $0.1$ and is updated by cosine learning rate scheduler until 210 epochs.
For other training parameters, we follow the implementation details in PyTorch.
For hyperparameters of DCKD, we use $\beta_{CE}$ as $1$, $\beta_{KD}$ as $1$ and $\beta_{Col}$ as $0.2$. The temperature of the knowledge distillation loss $T_{KD}$ is set to $4$, and the temperature of the collection loss $T_{KLD}$ is set to $2$.

\begin{table*}[h!]
	\centering
		\resizebox{\textwidth}{!}{
	\begin{tabular}{|l|l|cc|ccccccc|ccc|}
		\hline
		
		&&\textbf{Teacher} & \textbf{Student} & \multicolumn{10}{c|}{\textbf{Method}} \\
		\hline
		\multirow{2}{*}{Trial \#}&\multirow{2}{*}{\textbf{Accuracy}}& \multirow{2}{*}{ResNet34} & \multirow{2}{*}{ResNet18}    & \multirow{2}{*}{KD}& \multirow{2}{*}{AT}& \multirow{2}{*}{CC}&
		\multirow{2}{*}{OFD} & \multirow{2}{*}{CRD}    & \multirow{2}{*}{CRD+KD}  & \multirow{2}{*}{ReviewKD}  & \multicolumn{3}{c|}{\textbf{DCKD}}\\
		&&&&&&&&&&&\textbf{Net1} &\textbf{Net2} &\textbf{Net3}\\
		\Xhline{3\arrayrulewidth}
		1&\textbf{Top-1} & 73.31 & 69.75 & 70.05 & 71.74 & 71.46 & 71.37 & 71.97 & 72.02 &72.31& 72.35 &72.16&72.14 \\
		&\textbf{Top-5} & 91.42 & 89.07 & 90.04 & 90.67 & 90.23 & 90.24 & 90.66 & 90.87 &90.77& 90.88 &90.99&90.93 \\
		\hline
		2&\textbf{Top-1} & 73.31 & 69.75 & 69.92 & 71.56 & 71.35 & 71.30 & 71.81 & 72.02 &72.24& 72.27&72.16&72.12 \\
		&\textbf{Top-5} & 91.42 & 89.07 & 89.93 & 90.34 & 90.22 & 90.27 & 90.50  & 90.80  &90.86& 90.96&90.92&90.92 \\
		\hline
		3&\textbf{Top-1} & 73.31 & 69.75 & 70.20  & 71.36 & 71.54 & 71.37 & 71.89 & 72.05 &72.19& 72.19&72.02&72.00 \\
		&\textbf{Top-5} & 91.42 & 89.07 & 90.01 & 90.37 & 90.17 & 90.33 & 90.61 & 90.88 &90.95& 90.95&90.95&90.99  \\
		\hline
		
	\end{tabular}%
	}
	\caption{Top1 and Top5 accuracies (\%) of compression methods on ImageNet validation set.
	}
	\label{tab:imagenet_tab}%
\end{table*}%

\clearpage
\subsection{CIFAR-100}
 For all experiments, we use a single GPU for training with batch size of 64. 
 The initial learning rate for MobileNet and ShuffleNet is set to $0.01$; otherwise, it is set to $0.05$.
 The learning rate is updated by the cosine annealing learning rate scheduler until 930 epochs. 
 For hyperparameters of DCKD, we use $\beta_{CE}$ as $1$, $\beta_{KD}$ as $1$ and $\beta_{Col}$ as $0.5$. The temperature of the knowledge distillation loss $T_{KD}$ is set to $4$, and the temperature of the collection loss $T_{KLD}$ is set to $2$.
\begin{table*}[h!]
	\centering
	\begin{tabular}{|c|c|c|c|c|c|c|c|}
		\hline
		\bfseries\makecell{Teacher Network\\(Accuracy)}
		& \makecell{WRN-40-2\\(76.35)}
		& \makecell{WRN-40-2\\(76.35)}
		& \makecell{ResNet56\\(73.23)}
		& \makecell{ResNet110\\(74.36)} 
		& \makecell{ResNet110\\(74.36)}
		& \makecell{ResNet32x4\\(79.54)}
		& \makecell{VGG13\\(74.93)}\\
		\bfseries\makecell{Student Network\\(Accuracy)} 
		& \makecell{WRN-16-2\\(73.32)} 
		& \makecell{WRN-40-1\\(71.83)} 
		& \makecell{ResNet20\\(69.28)} 
		& \makecell{ResNet20\\(69.28)} 
		& \makecell{ResNet32\\(71.50)} 
		& \makecell{ResNet8x4\\(72.83)} 
		& \makecell{VGG8\\(70.92)} \\
		\hhline{|========|}
		\textbf{Method} \footnotesize(Trial \#) & \multicolumn{7}{c|}{\textbf{Accuracy}} \\
		\Xhline{3\arrayrulewidth}
		KD \footnotesize(\#1)    & 75.78 & 74.69 & 71.38 & 71.91 & 74.12 & 73.40  & 73.67 \\
		KD \footnotesize(\#2)    & 75.88 & 74.36 & 71.87 & 72.40  & 74.26 & 73.88 & 73.79 \\
		KD \footnotesize(\#3)    & 75.67 & 74.80  & 72.44 & 72.11 & 74.35 & 73.20  & 73.96 \\
		\hline
		FitNet \footnotesize(\#1) & 73.55 & 72.38 & 69.42 & 69.20  & 70.85 & 73.31 & 71.27 \\
		FitNet \footnotesize(\#2) & 74.07 & 72.55 & 68.92 & 69.60  & 70.57 & 73.33 & 71.31 \\
		FitNet \footnotesize(\#3) & 73.54 & 72.50  & 68.92 & 69.43 & 70.76 & 73.08 & 71.48 \\
		\hline
		AT \footnotesize(\#1)    & 74.17 & 73.41 & 70.88 & 71.05 & 72.88 & 73.84 & 71.53 \\
		AT \footnotesize(\#2)    & 74.46 & 73.44 & 70.94 & 71.02 & 72.46 & 73.81 & 72.01 \\
		AT \footnotesize(\#3)    & 74.07 & 73.40  & 70.87 & 70.80  & 73.05 & 73.94 & 72.14 \\
		\hline
		RKD \footnotesize(\#1)   & 74.41 & 72.76 & 70.67 & 70.12 & 72.66 & 71.95 & 71.34 \\
		RKD \footnotesize(\#2)   & 73.68 & 72.46 & 70.82 & 70.63 & 72.52 & 71.51 & 71.51 \\
		RKD \footnotesize(\#3)   & 73.82 & 72.89 & 71.27 & 70.87 & 72.78 & 71.78 & 71.46 \\
		\hline
		FT \footnotesize(\#1)    & 73.33 & 72.14 & 70.55 & 71.06 & 71.73 & 73.63 & 71.24 \\
		FT \footnotesize(\#2)    & 73.81 & 72.11 & 70.58 & 70.75 & 72.87 & 73.49 & 71.66 \\
		FT \footnotesize(\#3)    & 73.94 & 72.18 & 71.19 & 70.92 & 72.49 & 73.47 & 70.96 \\
		\hline
		CC \footnotesize(\#1)    & 73.50  & 71.47 & 69.77 & 69.55 & 71.56 & 72.87 & 70.65 \\
		CC \footnotesize(\#2)    & 73.51 & 72.36 & 70.35 & 69.75 & 71.59 & 72.61 & 70.47 \\
		CC \footnotesize(\#3)    & 73.12 & 72.32 & 70.49 & 69.97 & 71.60  & 72.48 & 70.43 \\
		\hline
		OFD \footnotesize(\#1) & 74.95	&74.00	&71.34	&71.41	&73.05	&74.34	&71.00\\
		OFD \footnotesize(\#2) & 75.26	&74.33	&71.66	&71.63	&73.18	&74.45	&71.16\\
		OFD \footnotesize(\#3) & 75.74	&74.30	&71.43	&71.73	&73.52	&74.25	&71.24\\
		\hline
		
		CRD \footnotesize(\#1)   & 76.16 & 74.54 & 72.73 & 72.12 & 74.05 & 75.55 & 73.79 \\
		CRD \footnotesize(\#2)   & 76.12 & 75.00    & 72.00    & 72.10  & 74.29 & 75.57 & 74.00 \\
		CRD \footnotesize(\#3)   & 75.95 & 74.51 & 72.43 & 72.61 & 74.24 & 75.70  & 73.62 \\
		\hline
		CRD+KD \footnotesize(\#1) & 76.08 & 75.35 & 72.69 & 72.63 & 74.64 & 75.61 & 74.40 \\
		CRD+KD \footnotesize(\#2) & 76.23 & 75.47 & 72.53 & 72.56 & 74.93 & 75.50  & 74.28 \\
		CRD+KD \footnotesize(\#3) & 75.93 & 75.68 & 72.80  & 72.78 & 74.60  & 75.37 & 74.44 \\    
		
		\hline
		ReviewKD \footnotesize(\#1) &75.82&	74.82&	71.80&	71.98&	73.60&	74.39&	72.02\\
		ReviewKD \footnotesize(\#2) & 75.77&	74.77&	71.99&	72.25&	73.61&	73.88&	72.00\\
		ReviewKD \footnotesize(\#3) & 75.65&	75.12&	72.00&	72.04&	73.82&	74.27&	72.01\\  
		\Xhline{3\arrayrulewidth}
		\textbf{DCKD Net1}\footnotesize(\#1)	&76.77	&75.98	&72.90	&73.09	&75.39	&75.79	&75.16\\
		\textbf{DCKD Net2}\footnotesize(\#1)	&76.54	&75.62	&72.65	&72.81	&75.03	&75.78	&75.09\\
		\textbf{DCKD Net3}\footnotesize(\#1)	&76.39	&75.44	&72.54	&72.68	&74.81	&75.34	&74.76\\
		\textbf{DCKD Net1}\footnotesize(\#2)	&76.93	&75.95	&73.27	&72.97	&75.27	&76.02	&74.74\\
		\textbf{DCKD Net2}\footnotesize(\#2)	&76.57	&75.68	&72.86	&72.82	&75.18	&75.82	&74.66\\
		\textbf{DCKD Net3}\footnotesize(\#2)	&76.55	&75.63	&72.43	&72.66	&75.06	&75.64	&74.63\\
		\textbf{DCKD Net1}\footnotesize(\#3)	&76.86	&75.75	&73.10	&73.14	&75.32	&76.18	&74.80\\
		\textbf{DCKD Net2}\footnotesize(\#3)	&76.74	&75.65	&72.91	&73.00	&75.10	&75.92	&74.53\\
		\textbf{DCKD Net3}\footnotesize(\#3)	&76.63	&75.62	&72.85	&72.75	&74.75	&75.81	&74.34\\
		\hline
		\end{tabular}%
		\caption{Top1 accuracies (\%) of student models on CIFAR-100 validation set that have architectures that are similar to their teacher models. 
		}
		\label{tab:cifar1}%
		\end{table*}%

\begin{table*}[h!]
  \centering
    \begin{tabular}{|c|c|c|c|c|c|c|}
    \hline
    \bfseries\makecell{Teacher Network\\(Accuracy)}
    & \makecell{VGG13\\(74.93)}
    & \makecell{ResNet50\\(79.68)}
    & \makecell{ResNet50\\(79.68)}
    & \makecell{ResNet32x4\\(79.54)} 
    & \makecell{ResNet32x4\\(79.54)}
    & \makecell{WRN-40-2    \\(76.35)} \\
    \bfseries\makecell{Student Network\\(Accuracy)} 
    & \makecell{MobileNetV2\\(65.23)} 
    & \makecell{MobileNetV2\\(65.23)} 
    & \makecell{VGG8\\(70.92)} 
    & \makecell{ShuffleNetV1\\(71.34)} 
    & \makecell{ShuffleNetV2\\(73.89)} 
    & \makecell{ShuffleNetV1\\(71.34)} \\
    \hhline{|=======|}
    \textbf{Method} \footnotesize(Trial \#) & \multicolumn{6}{c|}{\textbf{Accuracy}} \\
    \Xhline{3\arrayrulewidth}
    KD \footnotesize(\#1)    & 69.01 & 69.50 & 74.21 & 74.97 & 76.26 & 76.29 \\
    KD \footnotesize(\#2)    & 69.89 & 69.10 & 74.35 & 75.34 & 75.70 & 76.99 \\
    KD \footnotesize(\#3)    & 69.02 & 69.78 & 74.23 & 75.99 & 75.44 & 76.62 \\
    \hline
    FitNet \footnotesize(\#1) & 63.39 & 62.45 & 69.48 & 74.12 & 74.00 & 74.59 \\
    FitNet \footnotesize(\#2) & 64.55 & 62.00 & 69.38 & 73.38 & 74.27 & 74.17 \\
    FitNet \footnotesize(\#3) & 62.30 & 61.88 & 69.52 & 73.71 & 74.19 & 74.19 \\
    \hline
    AT \footnotesize(\#1)    & 64.35 & 64.09 & 72.56 & 74.41 & 74.30 & 75.09 \\
    AT \footnotesize(\#2)    & 64.19 & 63.35 & 72.19 & 74.61 & 74.16 & 74.94 \\
    AT \footnotesize(\#3)    & 64.48 & 63.75 & 72.57 & 74.17 & 74.39 & 74.96 \\
    \hline
    RKD \footnotesize(\#1)   & 64.87 & 65.47 & 71.45 & 74.06 & 75.20 & 75.24 \\
    RKD \footnotesize(\#2)   & 66.16 & 66.32 & 71.77 & 74.45 & 74.94 & 75.05 \\
    RKD \footnotesize(\#3)   & 65.63 & 65.85 & 71.85 & 74.00 & 75.35 & 75.07 \\
    \hline
    FT \footnotesize(\#1)    & 65.97 & 65.67 & 71.65 & 74.44 & 75.10 & 74.96 \\
    FT \footnotesize(\#2)    & 65.22 & 64.86 & 71.88 & 74.64 & 74.94 & 75.26 \\
    FT \footnotesize(\#3)    & 65.90 & 65.51 & 71.80 & 74.60 & 74.89 & 74.50 \\
    \hline
    CC \footnotesize(\#1)    & 64.98 & 65.49 & 70.37 & 72.68 & 74.27 & 72.53 \\
    CC \footnotesize(\#2)    & 64.63 & 65.20 & 70.62 & 71.98 & 73.38 & 73.22 \\
    CC \footnotesize(\#3)    & 64.63 & 65.27 & 70.11 & 72.77 & 73.77 & 72.49 \\
    \hline
    OFD \footnotesize(\#1) &62.80	&63.46	&71.81	&75.40	&76.58	&76.20\\
    OFD \footnotesize(\#2) &64.47	&63.41	&71.34	&75.91	&76.91	&75.73\\
    OFD \footnotesize(\#3) &63.03	&63.08	&72.23	&75.38	&76.95	&75.90\\
    \hline
    CRD \footnotesize(\#1)   & 68.75 & 69.41 & 74.13 & 75.95 & 76.09 & 76.74 \\
    CRD \footnotesize(\#2)   & 69.66 & 69.19 & 74.08 & 76.14 & 76.02 & 76.51 \\
    CRD \footnotesize(\#3)   & 69.74 & 69.05 & 74.09 & 75.93 & 76.08 & 76.74 \\
    \hline
    CRD+KD \footnotesize(\#1) & 70.50 & 71.18 & 74.56 & 76.40 & 76.95 & 76.97 \\
    CRD+KD \footnotesize(\#2) & 70.67 & 71.02 & 74.76 & 76.79 & 77.57 & 77.48 \\
    CRD+KD \footnotesize(\#3) & 70.53 & 70.98 & 74.48 & 76.08 & 77.03 & 77.89 \\
    \hline
	ReviewKD\footnotesize(\#1) &67.12	&66.40	&70.98	&75.92	&76.46	&76.94\\
	ReviewKD\footnotesize(\#2) &68.10	&66.76	&70.52	&75.60	&75.91	&77.07\\
	ReviewKD\footnotesize(\#3) &67.51	&66.69	&71.08	&76.03	&75.85	&76.81\\
	
    \Xhline{3\arrayrulewidth}
    \textbf{DCKD Net1} \footnotesize(\#1) &71.13	&71.26	&75.02	&76.67	&77.62	&77.99\\
    \textbf{DCKD Net2} \footnotesize(\#1) &70.67	&70.45	&74.76	&76.63	&77.56	&77.29\\
    \textbf{DCKD Net3} \footnotesize(\#1) &70.60	&70.29	&74.59	&76.36	&77.22	&77.20\\
    \textbf{DCKD Net1} \footnotesize(\#2) &70.95	&71.42	&75.09	&76.71	&77.55	&77.82\\
    \textbf{DCKD Net2} \footnotesize(\#2) &70.72	&71.16	&74.86	&76.35	&77.40	&77.29\\
    \textbf{DCKD Net3} \footnotesize(\#2) &70.63	&71.08	&74.47	&76.35	&77.14	&77.09\\
    \textbf{DCKD Net1} \footnotesize(\#3) &70.99	&71.53	&75.05	&76.86	&77.62	&77.85\\
    \textbf{DCKD Net2} \footnotesize(\#3) &70.94	&71.31	&74.62	&76.44	&77.52	&77.65\\
    \textbf{DCKD Net3} \footnotesize(\#3) &70.79	&70.81	&74.61	&76.31	&77.41	&77.47\\
    
	\hline        
    \end{tabular}%
  \caption{Top1 accuracies (\%) of student models on CIFAR-100 validation set that have architectures that are different from their teacher models. 
  }
  \label{tab:cifar2}%
\end{table*}%

\newcolumntype{?}{!{\vrule width 1pt}}
\begin{table}[h!]
	\centering
	\begin{tabular}{|l|l|ccc|}
		\hline
		Trial \# & \bfseries{Method} &    \bfseries\makecell{Student\\Network1}    
		& \bfseries\makecell{Student\\Network2} & \bfseries\makecell{Student\\Network3}\\
		\Xhline{3\arrayrulewidth}
		1&DML	&74.50	&74.41	&74.36\\
		2&DML	&74.41	&74.26	&74.20\\
		3&DML	&74.88	&74.56	&74.41\\
		\hline
		1&ONE	&74.44	&74.30	&74.09\\
		2&ONE	&74.08	&73.75	&73.48\\
		3&ONE	&74.05	&73.95	&73.71\\
		\hline
		1&OKDDip	&74.94	&74.65	&74.48\\
		2&OKDDip	&74.83	&74.46	&74.41\\
		3&OKDDip	&74.85	&74.64	&74.44\\
		
		\hline
		1&KDCL	&73.41	&73.15	&72.95\\
		2&KDCL	&73.45	&73.23	&72.99\\
		3&KDCL	&73.39	&73.29	&73.21\\

		\Xhline{3\arrayrulewidth}
		1&\textbf{DCKD}	&75.39	&75.03	&74.81\\
		2&\textbf{DCKD}	&75.27	&75.18	&75.06\\
		3&\textbf{DCKD}	&75.32	&75.10	&74.75\\
		\hline
	\end{tabular}
	\caption{Validation accuracies (\%) of the multistudent methods with the teacher on CIFAR-100. The teacher network is ResNet110 and the student is ResNet32. The KD loss was added to each student in DML and OKDDip, except ONE. 
For KDCL, we used the pretrained teacher for a large network because KDCL is a method of training a large network and several small networks from scratch.
	}
	\label{tab:multstu}
\end{table}

\begin{table}[h!]
	\centering
	{\small
		\renewcommand{\tabcolsep}{1.8mm}
		\begin{tabular}{|l|l|cc|ccc|cccc|ccccc|}
			\hline
			Trial& \multicolumn{1}{l|}{\textbf{\# of}} & \multicolumn{2}{c|}{\textbf{2}} & \multicolumn{3}{c|}{\textbf{3}} & \multicolumn{4}{c|}{\textbf{4}}      & \multicolumn{5}{c|}{\textbf{5}} \\
			\# &\textbf{students}     
			& \textbf{Net1}  & \textbf{Net2}  & \textbf{Net1}  & \textbf{Net2}  & \textbf{Net3}  & \textbf{Net1}  & \textbf{Net2}  & \textbf{Net3}  & \textbf{Net4}  & \textbf{Net1}  & \textbf{Net2}  & \textbf{Net3}  & \textbf{Net4}  & \textbf{Net5} \\
			\Xhline{3\arrayrulewidth}
			1& \textbf{DCKD}	&75.14	&74.86	&75.39	&75.03	&74.81	&75.08	&74.98	&74.90	&74.72	&75.02	&75.00	&74.99	&74.77	&74.76\\
			2& \textbf{DCKD}	&75.22	&74.90	&75.27	&75.18	&75.06	&75.01	&74.78	&74.75	&74.73	&75.03	&74.86	&74.81	&74.80	&74.64\\
			3& \textbf{DCKD}	&75.16	&74.78	&75.32	&75.10	&74.75	&75.08	&74.92	&74.77	&74.72	&75.09	&75.03	&74.97	&74.88	&74.52\\
			
			\hline
			
		\end{tabular}%
	}
	\label{tab:addlabel}%
	\caption{Validation accuracies (\%) of the DCKD on CIFAR-100 per the number of students. The teacher network is ResNet110 and the student is ResNet32. 
	}
\end{table}%

\textsc{\clearpage}

\begin{table}[h]
	\begin{center}
		\begin{tabular}{|l|l|ccc|}
			\hline
			Trial \# &\bfseries{Direction of KL Divergence} &	\bfseries\makecell{Student\\Network1}	
			& \bfseries\makecell{Student\\Network2} & \bfseries\makecell{Student\\Network3}\\
  		\Xhline{3\arrayrulewidth}
  		
  		1&DCKD with Reverse KL, Max collection method 	&75.39	&75.03	&74.81\\
  		2&DCKD with Reverse KL, Max collection method	&75.27	&75.18	&75.06\\
  		3&DCKD with Reverse KL, Max collection method	&75.32	&75.10	&74.75\\
  		\hline
  		1&DCKD with Forward KL, Max collection method	&75.03	&74.63	&74.60\\
  		2&DCKD with Forward KL, Max collection method	&75.06	&74.94	&74.63\\
  		3&DCKD with Forward KL, Max collection method	&74.93	&74.59	&74.49\\
  		\hline
  		1&DCKD with Bidirectional KL, Max collection method	&74.84	&74.71	&74.51\\
  		2&DCKD with Bidirectional KL, Max collection method	&74.97	&74.82	&74.79\\
  		3&DCKD with Bidirectional KL, Max collection method	&74.87	&74.84	&74.79\\
  		\hline
  		1&DCKD with Reverse KL, Average collection method	&75.09	&74.95	&74.82\\
  		2&DCKD with Reverse KL, Average collection method	&75.06	&75.02	&74.84\\
  		3&DCKD with Reverse KL, Average collection method	&75.01	&74.96	&74.76\\
  		\hline
  		1&DCKD with Forward KL, Average collection method	&74.88	&74.87	&74.82\\
  		2&DCKD with Forward KL, Average collection method	&74.87	&74.72	&74.66\\
  		3&DCKD with Forward KL, Average collection method	&74.91	&74.89	&74.59\\
  		\hline

		\end{tabular}
	\end{center}
	\caption{Validation accuracies (\%) of DCKD on CIFAR-100 for directions of KL divergence and collection method. The teacher network is ResNet110 and the student is ResNet32. 
	}
	\label{tab:kl_col}
\end{table}

\begin{table}[h]
	\centering
	\renewcommand{\tabcolsep}{1.5mm}
	\begin{tabular}{|c|c|c|c|c|c|c|c|}
		\hline
		\bfseries\makecell{Teacher of DCKD\\(Accuracy)}
		&\makecell{WRN-40-2\\(76.35)}
		&\makecell{WRN-40-2\\(76.35)}	
		&\makecell{ResNet56\\(73.23)}
		&\makecell{ResNet110\\(74.36)}
		&\makecell{ResNet110\\(74.36)}
		&\makecell{ResNet32x4\\(79.54)}
		&\makecell{VGG13\\(74.93)}\\
		\hline
		\bfseries\makecell{Teacher (Accuracy)\\DCKD Net1 \\DCKD Net2\\ DCKD Net3} 
		& \makecell{WRN-16-2\\(76.85)\\(76.62)\\(76.52)} 
		& \makecell{WRN-40-1\\(75.89)\\(75.65)\\(75.56)}
		& \makecell{ResNet20\\(73.09)\\(72.81)\\(72.61)} 
		& \makecell{ResNet20\\(73.07)\\(72.88)\\(72.70)} 
		& \makecell{ResNet32\\(75.33)\\(75.10)\\(74.87)} 
		& \makecell{ResNet8x4\\(76.00)\\(75.84)\\(75.60)} 		
		& \makecell{VGG8\\(74.90)\\(74.76)\\(74.58)}\\

		\bfseries\makecell{Student Network\\(Accuracy)} 
		& \makecell{WRN-16-2\\(73.32)} 
		& \makecell{WRN-40-1\\(71.83)}
		& \makecell{ResNet20\\(69.28)} 
		& \makecell{ResNet20\\(69.28)} 		
		& \makecell{ResNet32\\(71.50)} 
		& \makecell{ResNet8x4\\(72.83)} 
		& \makecell{VGG8\\(70.92)}\\
		
		\hhline{|========|}
		\textbf{Method} \footnotesize(Trial \#)& \multicolumn{7}{c|}{\textbf{Accuracy}} \\
		\Xhline{3\arrayrulewidth}
		\textbf{eDCKD Net1}	\footnotesize(\#1) &\textbf{77.09}	&\textbf{76.26}	&72.74	&72.53	&\textbf{75.58}	&75.95	&\textbf{75.03}\\
		\textbf{eDCKD Net2}	\footnotesize(\#1) &\textbf{77.05}	&\textbf{76.16}	&72.62	&72.45	&\textbf{75.46}	&75.85	&\textbf{74.97}\\
		\textbf{eDCKD Net3}	\footnotesize(\#1) &76.77	&\textbf{75.96}	&72.41	&72.15	&\textbf{75.45}	&75.85	&74.87\\
		\textbf{eDCKD Net1}	\footnotesize(\#2) &\textbf{77.29}	&\textbf{76.43}	&72.70	&72.64	&\textbf{75.52}	&\textbf{76.42}	&\textbf{75.18}\\
		\textbf{eDCKD Net2}	\footnotesize(\#2) &\textbf{77.06}	&\textbf{76.38}	&72.62	&72.64	&75.18	&\textbf{76.35}	&\textbf{75.16}\\
		\textbf{eDCKD Net3}	\footnotesize(\#2) &\textbf{76.89}	&\textbf{76.00}	&72.56	&72.55	&75.16	&\textbf{76.35}	&\textbf{75.02}\\
		\textbf{eDCKD Net1}	\footnotesize(\#3) &\textbf{77.18}	&\textbf{76.45}	&72.67	&72.90	&\textbf{75.53}	&\textbf{76.67}	&\textbf{75.57}\\
		\textbf{eDCKD Net2}	\footnotesize(\#3) &\textbf{77.04}	&\textbf{76.26}	&72.54	&72.71	&\textbf{75.34}	&\textbf{76.48}	&\textbf{75.14}\\
		\textbf{eDCKD Net3}	\footnotesize(\#3) &\textbf{77.03}	&75.88	&72.33	&72.30	&75.10	&\textbf{76.41}	&\textbf{75.08}\\
		\hline
		\textbf{Ensembled Student} \footnotesize(\#1)	&\textbf{77.09}	&\textbf{76.11}	&72.70	&72.91	&\textbf{75.53}	&\textbf{76.15}	&\textbf{75.42}\\
		\textbf{Ensembled Student} \footnotesize(\#2)	&\textbf{77.19}	&\textbf{76.34}	&72.54	&72.69	&\textbf{75.37}	&\textbf{76.12}	&\textbf{75.12}\\
		\textbf{Ensembled Student} \footnotesize(\#3)	&\textbf{76.88}	&75.87	&72.58	&72.81	&74.94	&\textbf{76.18}	&\textbf{75.27}\\
		
		\hline
	\end{tabular}%
	\caption{Top1 accuracies (\%) of various student models on CIFAR-100 validation set. 
		The teacher models are DCKD models in Table \ref{tab:cifar1}.
		We performed experiments to show that DCKD's student can be good teachers because they have rich knowledge.
		Several experiments, \eg WRN-16-2, WRN-40-1, ResNet32, ResNet8x4 and VGG8, showed that the enhanced DCKD (eDCKD) models and ensembled students performed better than DCKD's students. 
	}
	\label{tab:edckd1}%
\end{table}%

\begin{table}[h]
	\centering
	\renewcommand{\tabcolsep}{1.5mm}
	\begin{tabular}{|c|c|c|c|c|c|c|}
		\hline
		\bfseries\makecell{Teacher of DCKD\\(Accuracy)}
		&\makecell{VGG13\\(74.93)}
		&\makecell{ResNet50\\(79.68)}	
		&\makecell{ResNet50\\(79.68)}
		&\makecell{ResNet32x4\\(79.54)}
		&\makecell{ResNet32x4\\(79.54)}
		&\makecell{WRN-40-2\\(76.35)}\\
		\hline
		\bfseries\makecell{Teacher (Accuracy)\\DCKD Net1\\DCKD Net2\\DCKD Net3} 
		& \makecell{MobileNetV2\\(71.02)\\	(70.78)	\\(70.67)} 
		& \makecell{MobileNetV2\\(71.40)\\	(70.97)	\\(70.73)} 
		& \makecell{VGG8\\(75.05)\\	(74.75)	\\(74.56)} 
		& \makecell{ShuffleNetV1\\(76.75)\\	(76.47)	\\(76.34)} 
		& \makecell{ShuffleNetV2\\(77.60)\\	(77.49)	\\(77.26)} 
		& \makecell{ShuffleNetV1\\(77.89)\\	(77.41)	\\(77.25)} \\
		\bfseries\makecell{Student Network\\(Accuracy)} 
		& \makecell{MobileNetV2\\(65.23)} 
		& \makecell{MobileNetV2\\(65.23)} 
		& \makecell{VGG8\\(70.92)} 
		& \makecell{ShuffleNetV1\\(71.34)} 
		& \makecell{ShuffleNetV2\\(73.89)} 
		& \makecell{ShuffleNetV1\\(71.34)} \\
		
		\hhline{|=======|}
		\textbf{Method} \footnotesize(Trial \#)& \multicolumn{6}{c|}{\textbf{Accuracy}} \\
		\Xhline{3\arrayrulewidth}
		\textbf{eDCKD Net1} \footnotesize(\#1)	&\textbf{71.64}	&\textbf{71.46}	&74.81	&\textbf{76.82}	&77.46	&77.49\\
		\textbf{eDCKD Net2} \footnotesize(\#1)	&\textbf{71.11}	&71.23	&74.79	&76.41	&77.01	&77.36\\
		\textbf{eDCKD Net3} \footnotesize(\#1)	&\textbf{71.11}	&71.18	&74.63	&75.98	&76.82	&77.03\\
		\textbf{eDCKD Net1} \footnotesize(\#2)	&\textbf{72.01}	&\textbf{71.87}	&\textbf{75.17}	&76.41	&77.35	&77.57\\
		\textbf{eDCKD Net2} \footnotesize(\#2)	&\textbf{71.81}	&\textbf{71.80}	&\textbf{75.14}	&76.15	&77.34	&77.10\\
		\textbf{eDCKD Net3} \footnotesize(\#2)	&\textbf{71.76}	&\textbf{71.77}	&\textbf{75.08}	&76.10	&77.00	&77.05\\
		\textbf{eDCKD Net1} \footnotesize(\#3)	&\textbf{71.92}	&\textbf{71.92}	&75.00	&76.03	&77.31	&77.48\\
		\textbf{eDCKD Net2} \footnotesize(\#3)	&\textbf{71.80}	&\textbf{71.63}	&74.96	&76.01	&76.98	&77.38\\
		\textbf{eDCKD Net3} \footnotesize(\#3)	&\textbf{71.64}	&\textbf{71.54}	&74.86	&75.91	&76.96	&76.92\\
		\hline
		\textbf{Ensembled Student} \footnotesize(\#1)	&70.83	&70.47	&75.03	&\textbf{76.78}	&77.50	&\textbf{78.01}\\
		\textbf{Ensembled Student} \footnotesize(\#2)	&70.99	&\textbf{71.66}	&\textbf{75.22}	&76.26	&76.97	&\textbf{77.94}\\
		\textbf{Ensembled Student} \footnotesize(\#3)	&\textbf{71.41}	&\textbf{71.66}	&\textbf{75.11}	&\textbf{76.92}	&77.30	&77.66\\
		\hline
	\end{tabular}%
	\caption{Top1 accuracies (\%) of student models on CIFAR-100 validation set.
		The teacher models are DCKD models in Table \ref{tab:cifar2}.
		We performed experiments to show that DCKD's student can be good teachers because they have rich knowledge.
		A few experiments, \eg MobileNetV2, showed that the enhanced DCKD (eDCKD) models and ensembled students performed better than DCKD's students. 
	 }
	\label{tab:edckd2}%
\end{table}%

\clearpage
\subsection{CIFAR-10}
 For all experiments, we use a single GPU for training with batch size of 64. 
  The initial learning rate for MobileNet and ShuffleNet is set to $0.01$; otherwise, it is set to $0.05$.
 The learning rate is updated by the cosine annealing learning rate scheduler until 210 epochs. 
For hyperparameters of DCKD, we use $\beta_{CE}$ as $1$, $\beta_{KD}$ as $1$ and $\beta_{Col}$ as $0.5$.
The temperature of the knowledge distillation loss $T_{KD}$ is set to $4$, and the temperature of the collection loss $T_{KLD}$ is set to $2$.
\begin{table*}[ht]
	\centering
	\begin{tabular}{|c|c|c|c|c|c|c|c|}
		\hline
		\bfseries\makecell{Teacher Network\\(Accuracy)}
		& \makecell{WRN-40-2\\(93.57)}
		& \makecell{WRN-40-2\\(93.57)}
		& \makecell{ResNet56\\(91.82)}
		& \makecell{ResNet110\\(92.10)} 
		& \makecell{ResNet110\\(92.10)}
		& \makecell{ResNet32x4\\(94.58)}
		& \makecell{VGG13\\(92.87)}\\
		\bfseries\makecell{Student Network\\(Accuracy)} 
		& \makecell{WRN-16-2\\(92.28)} 
		& \makecell{WRN-40-1\\(91.70)} 
		& \makecell{ResNet20\\(90.80)} 
		& \makecell{ResNet20\\(90.80)} 
		& \makecell{ResNet32\\(91.56)} 
		& \makecell{ResNet8x4\\(91.42)} 
		& \makecell{VGG8\\(90.32)} \\
		\hhline{|========|}
		\textbf{Method} \footnotesize(Trial \#) & \multicolumn{7}{c|}{\textbf{Accuracy}} \\
		\Xhline{3\arrayrulewidth}
		KD \footnotesize(\#1)    & 94.36 & 93.95 & 92.78 & 93.32 & 94.02 & 93.65 & 93.23 \\
		KD \footnotesize(\#2)	& 94.13	& 93.71	& 93.11	& 93.06	& 93.48	& 93.85	& 93.10 \\
		KD \footnotesize(\#3)	& 94.27	& 93.80	& 92.88	& 93.04	& 93.74	& 93.84	& 93.12 \\
		\hline
		FitNet \footnotesize(\#1) & 93.60 & 93.46 & 91.14 & 91.09 & 91.50 & 92.18 & 91.89 \\
		FitNet \footnotesize(\#2)	& 93.83	& 93.70	& 91.19	& 90.75	& 91.38	& 92.46	& 92.02 \\
		FitNet \footnotesize(\#3)	& 93.69	& 93.65	& 91.47	& 90.90	& 91.57	& 92.03	& 92.18 \\
		\hline
		AT \footnotesize(\#1)    & 93.63 & 93.68 & 92.33 & 92.14 & 92.96 & 93.04 & 91.81 \\
		AT \footnotesize(\#2)	& 94.21	& 93.79	& 91.99	& 92.35	& 92.79	& 92.96	& 91.73 \\
		AT \footnotesize(\#3)	& 93.94	& 94.00	& 92.16	& 92.24	& 92.82	& 93.07	& 91.88 \\
		\hline
		RKD \footnotesize(\#1)   & 94.08 & 93.76 & 92.79 & 92.67 & 93.30 & 93.60 & 92.83 \\
		RKD \footnotesize(\#2)	& 94.03	& 93.76	& 92.80	& 92.78	& 93.33	& 93.50	& 92.97 \\
		RKD \footnotesize(\#3)	& 94.13	& 93.84	& 92.60	& 92.96	& 93.45	& 93.31	& 92.89 \\
		\hline
		FT \footnotesize(\#1)    & 93.28 & 93.15 & 92.01 & 91.63 & 92.32 & 92.94 & 92.14 \\
		FT \footnotesize(\#2)	& 93.47	& 93.28	& 92.05	& 91.84	& 92.20	& 92.84	& 91.86 \\
		FT \footnotesize(\#3)	& 93.57	& 93.47	& 91.75	& 92.16	& 92.43	& 92.81	& 91.83 \\
		\hline
		CC \footnotesize(\#1)    & 94.18 & 93.54 & 92.56 & 92.59 & 93.90 & 92.44 & 91.72 \\
		CC \footnotesize(\#2)	& 93.93	& 93.86	& 92.63	& 92.85	& 93.55	& 92.60	& 91.73 \\
		CC \footnotesize(\#3)	& 94.14	& 93.57	& 93.00	& 92.85	& 93.76	& 92.57	& 92.02 \\
		\hline
		CRD \footnotesize(\#1)   & 90.89 & 89.24 & 89.94 & 89.99 & 90.03 & 92.14 & 88.42 \\
		CRD \footnotesize(\#2)	& 91.12	& 89.52	& 89.64	& 90.15	& 90.59	& 92.08	& 88.31 \\
		CRD \footnotesize(\#3)	& 91.63	& 89.50	& 89.71	& 90.20	& 90.44	& 91.74	& 88.62 \\
		\hline
		CRD+KD \footnotesize(\#1) & 92.96 & 92.38 & 91.89 & 92.57 & 92.91 & 93.64 & 90.51 \\
		CRD+KD \footnotesize(\#2)	& 93.24	& 92.20	& 92.01	& 92.19	& 93.08	& 93.52	& 90.90 \\
		CRD+KD \footnotesize(\#3)	& 92.82	& 92.38	& 91.67	& 92.42	& 92.88	& 93.66	& 90.49 \\
		\Xhline{3\arrayrulewidth}   
		\textbf{DCKD Net1} \footnotesize(\#1)	&94.60	&94.43	&93.27	&93.29	&94.07	&94.27	&93.28\\
		\textbf{DCKD Net2} \footnotesize(\#1)	&94.47	&94.21	&93.21	&93.25	&93.89	&94.01	&93.24\\
		\textbf{DCKD Net3} \footnotesize(\#1)	&94.45	&94.16	&93.19	&93.19	&93.87	&94.01	&92.92\\
		\textbf{DCKD Net1} \footnotesize(\#2)	&94.84	&94.27	&93.44	&93.52	&94.12	&94.15	&93.34\\
		\textbf{DCKD Net2} \footnotesize(\#2)	&94.78	&94.23	&93.35	&93.27	&93.85	&93.96	&93.28\\
		\textbf{DCKD Net3} \footnotesize(\#2)	&94.72	&93.90	&93.34	&93.16	&93.68	&93.65	&93.21\\
		\textbf{DCKD Net1} \footnotesize(\#3)	&94.66	&94.29	&93.34	&93.55	&94.07	&94.25	&93.42\\
		\textbf{DCKD Net2} \footnotesize(\#3)	&94.47	&94.25	&93.31	&93.29	&94.00	&94.00	&93.23\\
		\textbf{DCKD Net3} \footnotesize(\#3)	&94.44	&93.87	&93.18	&93.19	&93.91	&93.92	&93.18\\

		\hline
	\end{tabular}%
	\caption{Top1 accuracies (\%) of various student models on CIFAR-10 validation set that have architectures that are similar to their teacher models. 
	}
	\label{tab:cifar10}%
\end{table*}%

\begin{table*}[ht]
	\centering
	\begin{tabular}{|c|c|c|c|c|c|c|}
		\hline
		\bfseries\makecell{Teacher Network\\(Accuracy)}
		& \makecell{VGG13\\(92.87)}
		& \makecell{ResNet50\\(93.24)}
		& \makecell{ResNet50\\(93.24)}
		& \makecell{ResNet32x4\\(94.58)} 
		& \makecell{ResNet32x4\\(94.58)}
		& \makecell{WRN-40-2    \\(93.57)} \\
		\bfseries\makecell{Student Network\\(Accuracy)} 
		& \makecell{MobileNetV2\\(91.79)} 
		& \makecell{MobileNetV2\\(91.79)} 
		& \makecell{VGG8\\(90.32)} 
		& \makecell{ShuffleNetV1\\(89.26)} 
		& \makecell{ShuffleNetV2\\(91.79)} 
		& \makecell{ShuffleNetV1\\(89.26)} \\
		\hhline{|=======|}
		\textbf{Method} \footnotesize(Trial \#) & \multicolumn{6}{c|}{\textbf{Accuracy}} \\
		\Xhline{3\arrayrulewidth}
		KD \footnotesize(\#1)    & 90.40 & 90.14 & 93.03 & 93.35 & 93.84 & 93.48 \\
		KD \footnotesize(\#2)	& 89.91	& 89.94	& 92.90	& 93.49	& 93.49	& 93.25 \\
		KD \footnotesize(\#3)	& 90.25	& 90.14	& 93.17	& 93.18	& 93.50	& 93.27 \\
		\hline
		FitNet \footnotesize(\#1) & 89.25 & 88.07 & 91.16 & 93.25 & 93.78 & 93.13 \\
		FitNet \footnotesize(\#2)	& 89.39	& 87.71	& 90.72	& 93.28	& 93.55	& 93.40 \\
		FitNet \footnotesize(\#3)	& 89.22	& 88.18	& 90.83	& 93.59	& 93.74	& 93.23 \\
		\hline
		AT \footnotesize(\#1)    & 87.70 & 86.77 & 92.16 & 93.73 & 93.84 & 93.42 \\
		AT \footnotesize(\#2)	& 88.49	& 87.23	& 92.39	& 94.06	& 93.92	& 93.54 \\
		AT \footnotesize(\#3)	& 88.03	& 87.14	& 92.10	& 93.90	& 94.08	& 93.71 \\
		\hline
		RKD \footnotesize(\#1)   & 90.08 & 90.16 & 93.07 & 93.36 & 93.94 & 93.36 \\
		RKD \footnotesize(\#2)	& 90.34	& 90.38	& 93.10	& 93.50	& 93.90	& 93.18 \\
		RKD \footnotesize(\#3)	& 90.09	& 90.26	& 92.96	& 93.61	& 93.66	& 93.43 \\
		\hline
		FT \footnotesize(\#1)    & 87.52 & 85.90 & 92.17 & 93.17 & 93.61 & 92.85 \\
		FT \footnotesize(\#2)	& 88.22	& 87.05	& 92.19	& 93.23	& 93.32	& 92.88 \\
		FT \footnotesize(\#3)	& 88.42	& 87.69	& 92.27	& 93.80	& 93.73	& 93.55 \\
		\hline
		CC \footnotesize(\#1)    & 90.01 & 89.60 & 91.60 & 92.63 & 93.21 & 92.57 \\
		CC \footnotesize(\#2)	& 90.03	& 89.43	& 91.54	& 92.42	& 93.38	& 92.67 \\
		CC \footnotesize(\#3)	& 90.26	& 89.89	& 91.26	& 93.23	& 93.63	& 93.59 \\
		\hline
		CRD \footnotesize(\#1)   & 86.84 & 87.95 & 89.67 & 91.92 & 92.18 & 91.96 \\
		CRD \footnotesize(\#2)	& 86.35	& 88.01	& 90.03	& 91.84	& 92.08	& 91.70 \\
		CRD \footnotesize(\#3)	& 86.60	& 87.97	& 89.77	& 92.22	& 92.06	& 91.53 \\
		\hline
		CRD+KD \footnotesize(\#1) & 88.68 & 89.83 & 91.96 & 92.97 & 93.01 & 92.72 \\
		CRD+KD \footnotesize(\#2)	& 88.70	& 89.98	& 92.13	& 92.63	& 93.05	& 92.69 \\
		CRD+KD \footnotesize(\#3)	& 88.74	& 89.74	& 92.14	& 92.77	& 92.96	& 92.39 \\
		\hline
		\Xhline{3\arrayrulewidth}
		\textbf{DCKD Net1} \footnotesize(\#1)	&90.45	&90.54	&93.16	&94.15	&94.16	&94.13\\
		\textbf{DCKD Net2} \footnotesize(\#1)	&90.44	&90.53	&93.10	&94.08	&94.07	&93.97\\
		\textbf{DCKD Net3} \footnotesize(\#1)	&90.39	&90.45	&92.71	&93.92	&93.91	&93.88\\
		\textbf{DCKD Net1} \footnotesize(\#2)	&90.51	&90.50	&93.17	&94.22	&94.32	&93.95\\
		\textbf{DCKD Net2} \footnotesize(\#2)	&90.29	&90.27	&93.04	&94.15	&94.20	&93.73\\
		\textbf{DCKD Net3} \footnotesize(\#2)	&90.19	&90.22	&93.01	&94.04	&94.12	&93.70\\
		\textbf{DCKD Net1} \footnotesize(\#3)	&90.32	&90.61	&93.25	&94.25	&94.31	&94.16\\
		\textbf{DCKD Net2} \footnotesize(\#3)	&90.28	&90.55	&93.20	&94.21	&94.24	&93.86\\
		\textbf{DCKD Net3} \footnotesize(\#3)	&90.19	&90.03	&93.19	&94.07	&94.18	&93.68\\
		
		\hline
%
	\end{tabular}%
	\caption{Top1 accuracies (\%) of student models on CIFAR-10 validation set that have architectures that are different from their teacher models. In DCKD, we use $\beta_{CE}$ as $0.1$, $\beta_{KD}$ as $0.9$, $\beta_{Col}$ as $0.2$ and $T_{KLD}$ as $2$ for ShuffleNet. 
	}
	\label{tab:cifar102}%
\end{table*}%
\clearpage

\subsection{Fashion-MNIST}
 For all experiments, we use a single GPU for training with batch size of 64. 
The initial learning rate for MobileNet and ShuffleNet is set to $0.01$; otherwise, it is set to $0.05$.
The learning rate is updated by the cosine annealing learning rate scheduler until 450 epochs. 
For hyperparameters of DCKD, we use $\beta_{CE}$ as $1$, $\beta_{KD}$ as $1$ and $\beta_{Col}$ as $0.5$.
The temperature of the knowledge distillation loss $T_{KD}$ is set to $4$, and the temperature of the collection loss $T_{KLD}$ is set to $2$.
\begin{table*}[h!]
  \centering
    \begin{tabular}{|c|c|c|c|c|c|c|c|}
    \hline
    \bfseries\makecell{Teacher Network\\(Accuracy)}
    & \makecell{WRN-40-2\\(93.82)}
    & \makecell{WRN-40-2\\(93.82)}
    & \makecell{ResNet56\\(93.79)}
    & \makecell{ResNet110\\(94.08)} 
    & \makecell{ResNet110\\(94.08)}
    & \makecell{ResNet32x4\\(94.01)}
    & \makecell{VGG13\\(94.20)}\\
    \bfseries\makecell{Student Network\\(Accuracy)} 
    & \makecell{WRN-16-2\\(93.68)} 
    & \makecell{WRN-40-1\\(93.58)} 
    & \makecell{ResNet20\\(93.22)} 
    & \makecell{ResNet20\\(93.22)} 
    & \makecell{ResNet32\\(93.63)} 
    & \makecell{ResNet8x4\\(93.22)} 
    & \makecell{VGG8\\(93.45)} \\
    \hhline{|========|}
    \textbf{Method} \footnotesize(Trial \#) & \multicolumn{7}{c|}{\textbf{Accuracy}} \\
    \Xhline{3\arrayrulewidth}
    KD \footnotesize(\#1)    & 94.32 & 94.41 & 94.19 & 94.19 & 94.35 & 94.26 & 93.99 \\
    KD \footnotesize(\#2)	& 94.47	& 94.18	& 94.23	& 94.30	& 94.23	& 94.15	& 93.94 \\
    KD \footnotesize(\#3)	& 94.33	& 94.36	& 94.24	& 94.39	& 94.32	& 94.17	& 94.04 \\
    \hline
    FitNet \footnotesize(\#1) & 94.13 & 94.10 & 93.44 & 93.96 & 93.81 & 93.90 & 93.93 \\
    FitNet \footnotesize(\#2)	& 94.07	& 94.10	& 93.56	& 93.95	& 93.91	& 93.69	& 93.87 \\
    FitNet \footnotesize(\#3)	& 94.22	& 93.99	& 93.59	& 94.00	& 93.89	& 93.84	& 93.86 \\
    \hline
    AT \footnotesize(\#1)    & 93.91 & 94.07 & 93.65 & 94.16 & 94.13 & 93.59 & 93.55 \\
    AT \footnotesize(\#2)	& 94.14	& 93.98	& 93.64	& 94.07	& 94.14	& 93.92	& 93.56 \\
    AT \footnotesize(\#3)	& 93.96	& 94.01	& 93.79	& 94.21	& 94.37	& 93.80	& 93.59 \\
    \hline
    RKD \footnotesize(\#1)   & 94.00 & 93.95 & 93.94 & 94.28 & 94.30 & 94.05 & 93.82 \\
    RKD \footnotesize(\#2)	& 94.15	& 93.98	& 94.02	& 94.20	& 94.31	& 93.93	& 93.89 \\
    RKD \footnotesize(\#3)	& 94.06	& 94.27	& 93.96	& 94.25	& 94.35	& 94.12	& 94.01 \\
    \hline
    FT \footnotesize(\#1)    & 93.67 & 93.45 & 93.84 & 94.07 & 94.11 & 93.17 & 93.75 \\
    FT \footnotesize(\#2)	& 93.68	& 93.60	& 93.54	& 94.02	& 93.98	& 93.35	& 93.51 \\
    FT \footnotesize(\#3)	& 93.79	& 93.66	& 93.71	& 93.99	& 94.08	& 93.26	& 93.62 \\
    \hline
    CC \footnotesize(\#1)    & 94.07 & 94.13 & 94.11 & 94.18 & 94.13 & 93.75 & 93.65 \\
    CC \footnotesize(\#2)	& 93.94	& 93.94	& 94.11	& 94.05	& 94.10	& 93.72	& 93.79 \\
    CC \footnotesize(\#3)	& 94.07	& 94.26	& 93.93	& 94.12	& 94.41	& 93.62	& 93.75 \\
    \hline
    CRD \footnotesize(\#1)   & 91.89 & 91.27 & 91.65 & 91.58 & 91.88 & 93.04 & 91.63 \\
    CRD \footnotesize(\#2)	& 92.15	& 91.95	& 91.79	& 91.43	& 91.58	& 92.95	& 91.89 \\
    CRD \footnotesize(\#3)	& 92.04	& 90.98	& 90.75	& 91.33	& 91.59	& 93.01	& 92.04 \\
    \hline
    CRD+KD \footnotesize(\#1) & 93.45 & 93.14 & 93.20 & 92.94 & 93.16 & 93.83 & 92.98 \\
    CRD+KD \footnotesize(\#2)	& 93.47	& 92.90	& 93.20	& 92.83	& 93.39	& 93.83	& 92.72 \\
    CRD+KD \footnotesize(\#3)	& 93.67	& 93.04	& 93.36	& 93.15	& 93.27	& 93.91	& 92.83 \\
    \Xhline{3\arrayrulewidth}
    \textbf{DCKD Net1} \footnotesize(\#1)	&94.56	&94.51	&94.38	&94.61	&94.54	&94.35	&94.01\\
    \textbf{DCKD Net2} \footnotesize(\#1)	&94.44	&94.50	&94.37	&94.40	&94.53	&94.31	&93.97\\
    \textbf{DCKD Net3} \footnotesize(\#1)	&94.24	&94.34	&94.23	&94.29	&94.30	&94.19	&93.84\\
    \textbf{DCKD Net1} \footnotesize(\#2)	&94.44	&94.51	&94.58	&94.50	&94.56	&94.42	&94.09\\
    \textbf{DCKD Net2} \footnotesize(\#2)	&94.37	&94.38	&94.43	&94.50	&94.36	&94.24	&93.99\\
    \textbf{DCKD Net3} \footnotesize(\#2)	&94.31	&94.36	&94.38	&94.40	&94.29	&94.19	&93.94\\
    \textbf{DCKD Net1} \footnotesize(\#3)	&94.52	&94.51	&94.49	&94.35	&94.68	&94.36	&94.07\\
    \textbf{DCKD Net2} \footnotesize(\#3)	&94.40	&94.44	&94.41	&94.35	&94.52	&94.32	&94.06\\
    \textbf{DCKD Net3} \footnotesize(\#3)	&94.35	&94.26	&94.40	&94.33	&94.43	&94.21	&93.96\\
    
    \hline
%
    \end{tabular}%
  \caption{Top1 accuracies (\%) of various student models on Fashion-MNIST validation set that have architectures that are similar to their teacher models. 
  }
  \label{tab:fashin-mnist}%
\end{table*}%

\begin{table*}[ht]
  \centering
    \begin{tabular}{|c|c|c|c|c|c|c|}
    \hline
    \bfseries\makecell{Teacher Network\\(Accuracy)}
    & \makecell{VGG13\\(94.20)}
    & \makecell{ResNet50\\(93.41)}
    & \makecell{ResNet50\\(93.41)}
    & \makecell{ResNet32x4\\(94.01)} 
    & \makecell{ResNet32x4\\(94.01)}
    & \makecell{WRN-40-2    \\(93.82)} \\
    \bfseries\makecell{Student Network\\(Accuracy)} 
    & \makecell{MobileNetV2\\(93.12)} 
    & \makecell{MobileNetV2\\(93.12)} 
    & \makecell{VGG8\\(93.45)} 
    & \makecell{ShuffleNetV1\\(92.99)} 
    & \makecell{ShuffleNetV2\\(93.45)} 
    & \makecell{ShuffleNetV1\\(92.99)} \\
    \hhline{|=======|}
    \textbf{Method} \footnotesize(Trial \#) & \multicolumn{6}{c|}{\textbf{Accuracy}} \\
    \Xhline{3\arrayrulewidth}
    KD \footnotesize(\#1)    & 93.30 & 93.57 & 94.12 & 93.95 & 93.71 & 93.92 \\
    KD \footnotesize(\#2)	& 93.35	& 93.50	& 94.02	& 94.03	& 93.83	& 94.01 \\
    KD \footnotesize(\#3)	& 93.38	& 93.51	& 94.09	& 93.98	& 93.90	& 94.41 \\
    \hline
    FitNet \footnotesize(\#1) & 93.63 & 93.39 & 93.68 & 94.11 & 94.09 & 93.89 \\
    FitNet \footnotesize(\#2)	& 93.52	& 93.11	& 93.80	& 94.19	& 94.21	& 94.22 \\
    FitNet \footnotesize(\#3)	& 93.48	& 93.23	& 93.85	& 94.49	& 94.28	& 94.01 \\
    \hline
    AT \footnotesize(\#1)    & 92.96 & 92.62 & 93.63 & 94.16 & 94.05 & 94.02 \\
    AT \footnotesize(\#2)	& 93.21	& 92.61	& 93.76	& 94.12	& 94.08	& 94.20 \\
    AT \footnotesize(\#3)	& 93.33	& 92.81	& 93.69	& 94.18	& 94.28	& 94.09 \\
    \hline
    RKD \footnotesize(\#1)   & 93.40 & 93.52 & 94.12 & 94.11 & 93.76 & 93.91 \\
    RKD \footnotesize(\#2)	& 93.24	& 93.51	& 94.00	& 93.98	& 94.17	& 94.12 \\
    RKD \footnotesize(\#3)	& 93.15	& 93.60	& 93.94	& 94.09	& 93.82	& 94.21 \\
    \hline
    FT \footnotesize(\#1)    & 92.21 & 92.72 & 93.74 & 93.57 & 93.42 & 93.49 \\
    FT \footnotesize(\#2)	& 93.45	& 92.70	& 93.63	& 93.69	& 93.58	& 93.39 \\
    FT \footnotesize(\#3)	& 92.87	& 92.61	& 93.69	& 93.41	& 93.37	& 93.27 \\
    \hline
    CC \footnotesize(\#1)    & 93.49 & 93.65 & 93.83 & 94.22 & 93.60 & 93.98 \\
    CC \footnotesize(\#2)	& 93.46	& 93.53	& 93.77	& 93.85	& 93.63	& 93.96 \\
    CC \footnotesize(\#3)	& 93.56	& 93.16	& 93.84	& 94.08	& 93.58	& 94.06 \\
    \hline
    CRD \footnotesize(\#1)   & 89.74 & 90.30 & 91.93 & 92.78 & 93.90 & 92.80 \\
    CRD \footnotesize(\#2)	& 89.65	& 90.92	& 92.33	& 93.34	& 93.00	& 92.59 \\
    CRD \footnotesize(\#3)	& 89.28	& 90.29	& 92.29	& 92.93	& 92.76	& 92.69 \\
    \hline
    CRD+KD \footnotesize(\#1) & 91.19 & 92.29 & 93.44 & 93.16 & 93.54 & 93.69 \\
    CRD+KD \footnotesize(\#2) & 92.14	& 92.45	& 93.38	& 93.44	& 93.60	& 93.25 \\
    CRD+KD \footnotesize(\#3) & 91.69	& 92.40	& 93.24	& 93.43	& 93.28	& 93.42 \\
    \Xhline{3\arrayrulewidth}
    \textbf{DCKD Net1} (\#1)	&93.71	&93.67	&94.27	&94.84	&94.82	&94.78\\
    \textbf{DCKD Net2} (\#1)	&93.55	&93.64	&94.20	&94.66	&94.64	&94.58\\
    \textbf{DCKD Net3} (\#1)	&93.20	&93.57	&94.02	&94.92	&94.64	&94.64\\
    \textbf{DCKD Net1} (\#2)	&93.62	&93.77	&94.16	&94.83	&94.69	&94.53\\
    \textbf{DCKD Net2} (\#2)	&93.57	&93.51	&94.16	&94.83	&94.80	&94.65\\
    \textbf{DCKD Net3} (\#2)	&93.39	&93.20	&94.07	&94.77	&94.70	&94.57\\
    \textbf{DCKD Net1} (\#3)	&93.60	&93.80	&94.24	&94.77	&94.73	&94.55\\
    \textbf{DCKD Net2} (\#3)	&93.25	&93.66	&94.17	&94.88	&94.77	&94.73\\
    \textbf{DCKD Net3} (\#3)	&93.00	&93.34	&94.00	&94.78	&94.68	&94.50\\
    
    \hline
    \end{tabular}%
  \caption{Top1 accuracies (\%) of student models on Fashion-MNIST validation set that have architectures that are different from their teacher models. 
  }
  	
  \label{tab:fashin-mnist2}%
\end{table*}%

\clearpage

\subsection{The Number of Parameters for Network Architectures}
In this subsection, we report the number of parameters for teacher and student network architectures on each dataset. 
\begin{table}[h!]
	\centering
	\begin{tabular}{|cc|rr|}
		\hline
		\bfseries{Teacher} & \bfseries{Student} & \multicolumn{2}{c|}{\bfseries{Number of Parameters}}  \\
  		\Xhline{3\arrayrulewidth}
		ResNet34	&ResNet18		 & 21,797,672 	& 11,689,512   \\
		\hline
	\end{tabular}
	\caption{The number of parameters for teacher and student network architectures on ImageNet.}
\end{table}
\begin{table}[h!]
	\centering
	\begin{tabular}{|cc|rr|}
		\hline
		\bfseries{Teacher} & \bfseries{Student} & \multicolumn{2}{c|}{\bfseries{Number of Parameters}}  \\
  		\Xhline{3\arrayrulewidth}
		WRN-40-2	&WRN-16-2	& 2,255,156 &	 703,284  \\
		WRN-40-2	&WRN-40-1	& 2,255,156 &	 569,780  \\
		ResNet56	&ResNet20	& 861,620 	& 278,324     \\
		ResNet110	&ResNet20	& 1,736,564 &	 278,324  \\
		ResNet110	&ResNet32	& 1,736,564 &	 472,756  \\
		ResNet32x4	&ResNet8x4	& 7,433,860 &	 1,233,540\\
		VGG13		&VGG8	 	&9,462,180 	& 3,965,028   \\
		\hline			
		VGG13		&MobileNetV2	 &9,462,180 	& 812,836   \\
		ResNet50	&MobileNetV2	 &23,705,252 	& 812,836   \\
		ResNet50	&VGG8			 &23,705,252 	& 3,965,028 \\
		ResNet32x4	&ShuffleNetV1	 &7,433,860 	& 949,258   \\
		ResNet32x4	&ShuffleNetV2	 &7,433,860 	& 1,355,528 \\
		WRN-40-2	&ShuffleNetV1	 &2,255,156 	& 949,258   \\
		\hline
	\end{tabular}
	\caption{The number of parameters for teacher and student network architectures on CIFAR-100.}
\end{table}
\begin{table}[h!]
	\centering
	\begin{tabular}{|cc|rr|}
		\hline
		\bfseries{Teacher} & \bfseries{Student} & \multicolumn{2}{c|}{\bfseries{Number of Parameters}}  \\
  		\Xhline{3\arrayrulewidth}
		WRN-40-2	&WRN-16-2		 & 2,243,546 	& 691,674   \\
		WRN-40-2	&WRN-40-1		 & 2,243,546 	& 563,930   \\
		ResNet56	&ResNet20		 & 855,770 	 	&272,474    \\
		ResNet110	&ResNet20		 & 1,730,714 	& 272,474   \\
		ResNet110	&ResNet32		 & 1,730,714 	& 466,906   \\
		ResNet32x4	&ResNet8x4		 & 7,410,730 	& 1,210,410 \\
		VGG13		&VGG8	 		 &9,416,010 	& 3,918,858 \\
		\hline			                                       
		VGG13		&MobileNetV2	 &9,416,010 	& 697,546   \\
		ResNet50	&MobileNetV2	 & 23,520,842 	& 697,546   \\
		ResNet50	&VGG8			 &23,520,842 	& 3,918,858 \\
		ResNet32x4	&ShuffleNetV1	 &	 7,410,730 	& 862,768   \\
		ResNet32x4	&ShuffleNetV2	 &	 7,410,730 	& 1,263,278 \\
		WRN-40-2	&ShuffleNetV1	 &	 2,243,546 	& 862,768   \\
		\hline
	\end{tabular}
	\caption{The number of parameters for teacher and student network architectures on CIFAR-10 and Fashion-MNIST.}
\end{table}

\clearpage
\section{Figures}
We provide the additional examples for the visualization of correlation.

%
%
%

\begin{figure}[ht]
\centering
  \begin{subfigure}{0.5\linewidth}
  \centering
  \includegraphics[width=0.9\linewidth]{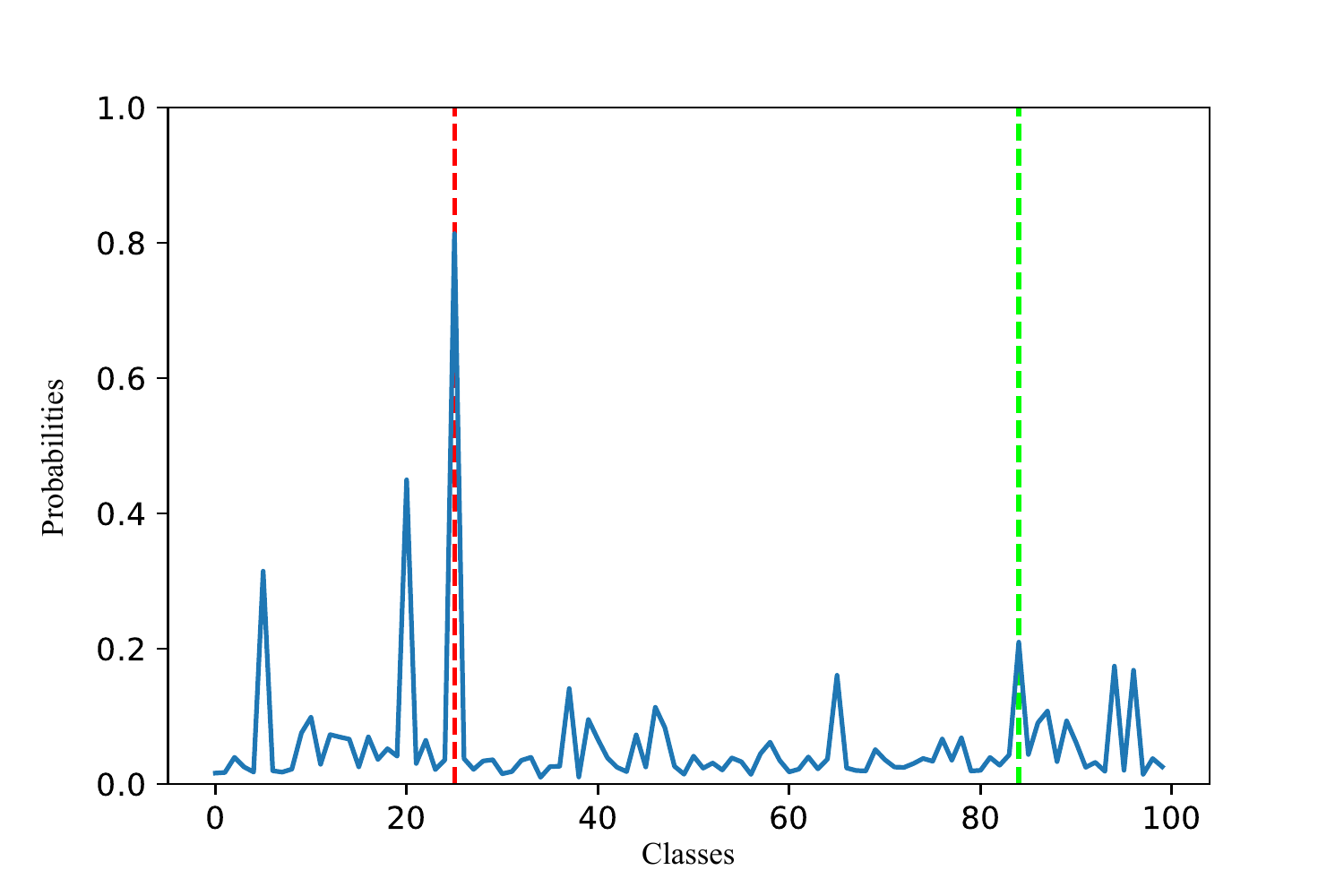}
    \caption{Accumulation of teacher outputs}
  \label{fig:mentor1}
  \end{subfigure}%
   \hspace*{-1.0cm}   
  \begin{subfigure}{0.5\linewidth}
  \centering
  \includegraphics[width=0.9\linewidth]{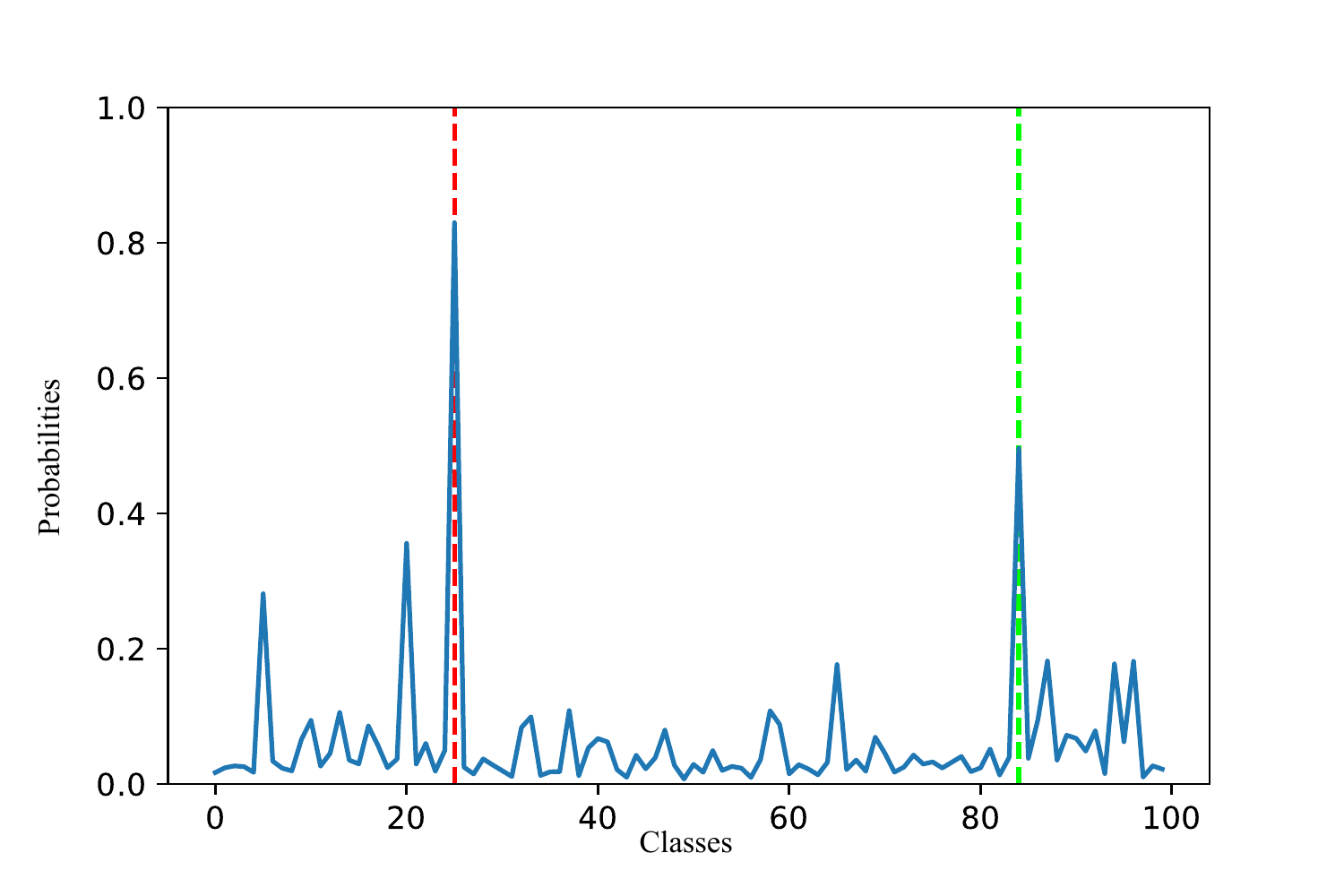}
     \caption{Accumulation of student outputs }
  \label{fig:mentee1}
  \end{subfigure}%
\caption{Accumulation of outputs of couch class images. The green line represents the table class (84th) and the red line represents the couch class (25th).
\cref{fig:mentor1} and \cref{fig:mentee1} are obtained by applying the maximum function to the outputs $\bf{p}$ of each of the teacher (ResNet110) and one of the DCKD's students (ResNet32) for all images of the couch class in CIFAR-100. 
 }\label{fig:1}
\end{figure}

\begin{figure}[ht]
\centering
  \begin{subfigure}{0.5\linewidth}
  \centering
  \includegraphics[width=1\linewidth]{figures/couch_25.eps}
    \caption{Class: Couch}
  \label{fig:couch}
  \end{subfigure}%
   \hspace*{-1.0cm}   
  \begin{subfigure}{0.5\linewidth}
  \centering
  \includegraphics[width=1\linewidth]{figures/table_84.eps}
     \caption{Class: Table}
  \label{fig:table}
  \end{subfigure}%
\caption{Sample images from couch and table classes. }\label{fig:2}
\end{figure}
\begin{figure}
\centering
  \begin{subfigure}{0.5\linewidth}
  \centering
  \includegraphics[width=0.9\linewidth]{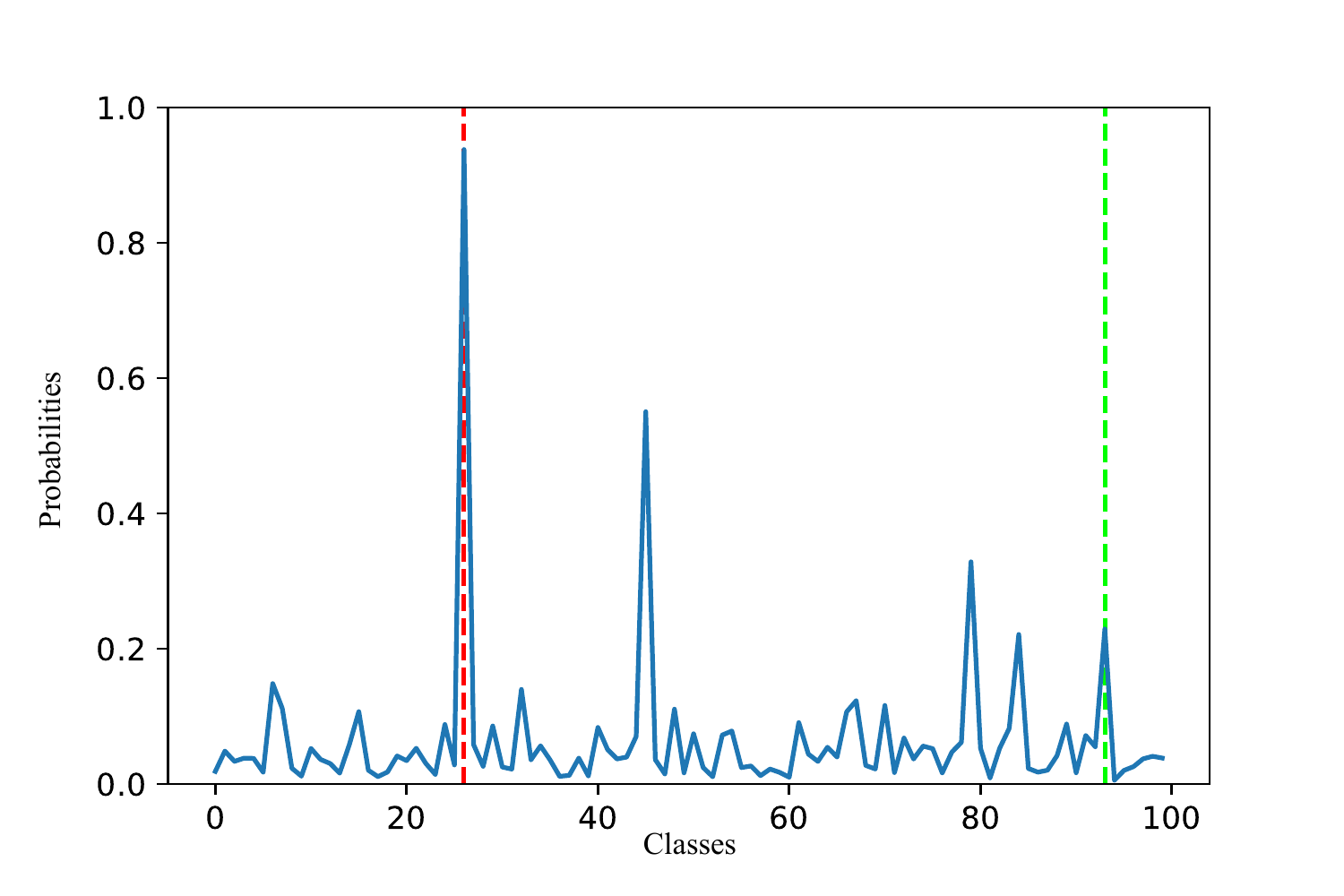}
    \caption{Accumulation of teacher outputs}
  \label{fig:mentor2}
  \end{subfigure}%
   \hspace*{-1.0cm}   
  \begin{subfigure}{0.5\linewidth}
  \centering
  \includegraphics[width=0.9\linewidth]{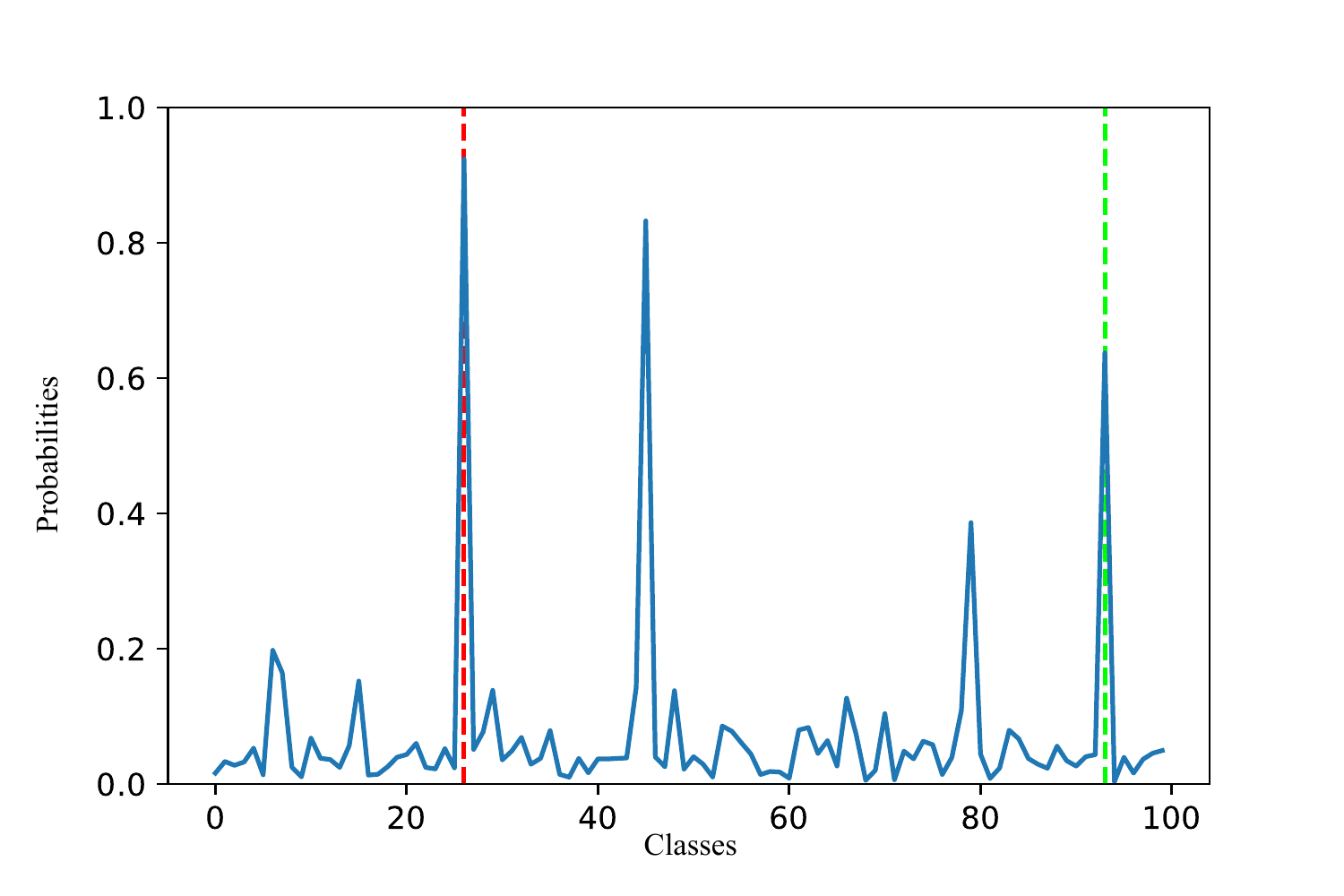}
     \caption{Accumulation of student outputs}
  \label{fig:mentee2}
  \end{subfigure}%
\caption{Accumulation of outputs of crab class images. The green line represents the turtle class (93rd) and the red line represents the crab class (26th).
\cref{fig:mentor2} and \cref{fig:mentee2} are obtained by applying the maximum function to the outputs $\bf{p}$ of each of the teacher (ResNet110) and one of the DCKD's students (ResNet32) for all images of the crab class in CIFAR-100. 
}\label{fig:3}
\end{figure}

\begin{figure}
\centering
  \begin{subfigure}{0.5\linewidth}
  \centering
  \includegraphics[width=1\linewidth]{figures/crab_26.eps}
    \caption{Class: Crab}
  \label{fig:crab}
  \end{subfigure}%
   \hspace*{-1.0cm}   
  \begin{subfigure}{0.5\linewidth}
  \centering
  \includegraphics[width=1\linewidth]{figures/turtle_93.eps}
     \caption{Class: Turtle}
  \label{fig:turtle}
  \end{subfigure}%
\caption{Sample images from crab and turtle classes. }\label{fig:4}
\end{figure}
%
%